\documentclass[smallcondensed]{svjour3}     
\smartqed  
\usepackage{subfigure}
\usepackage{graphicx}
\usepackage{mathrsfs}
\usepackage{stmaryrd}
\usepackage{pifont}
\usepackage{cite}
\usepackage{bbding}
\usepackage{bm}
\usepackage{multirow}
\usepackage{amssymb}
\usepackage{amssymb}
\usepackage{balance}
\usepackage{amsmath}
\usepackage{amsfonts}
\usepackage{amssymb}
\usepackage{algorithm}
\usepackage{multirow}
\usepackage{algpseudocode}
\usepackage{stfloats}
\usepackage{latexsym}
\usepackage{pdflscape}
\usepackage{float}
\usepackage{booktabs}
\usepackage{flushend}
\usepackage{comment}
\usepackage{array}
\usepackage{arydshln}
\usepackage{url}
\usepackage{color}
\usepackage{mathtools, nccmath}
\DeclarePairedDelimiter{\nint}\lfloor\rceil
\usepackage{pifont}
\usepackage[font=small,labelfont=bf]{caption} 

\journalname{Machine Learning (Springer)}
\begin{document}

\title{Multiscale Principle of Relevant Information for Hyperspectral Image Classification}


\author{Yantao Wei \and Shujian Yu \and \\
Luis Sanchez Giraldo \and Jos\'{e} C. Pr\'{i}ncipe}


\institute{
Yantao Wei \at
              Hubei Research Center for Educational Informationization, Central China Normal University, Wuhan 430079, China\\
              \email{yantaowei@mail.ccnu.edu.cn}\\
           \and
           Shujian Yu \at
           Department of Electrical and Computer Engineering, University of Florida, Gainesville, FL 32611, USA\\
           \email{yusjlcy9011@ufl.edu}
           \and
           Luis Sanchez Giraldo
           \at Department of Electrical and Computer Engineering, University of Kentucky, Lexington, KY 40508, USA\\
           \email{luis.sanchez@uky.edu}
           \and
           Jos\'{e} C. Pr\'{i}ncipe
           \at Department of Electrical and Computer Engineering, University of Florida, Gainesville, FL 32611, USA\\
           \email{principe@cnel.ufl.edu}}

\date{Received: date / Accepted: date}

\maketitle

\begin{abstract}
This paper proposes a novel architecture, termed multiscale principle of relevant information (MPRI), to learn discriminative spectral-spatial features for hyperspectral image (HSI) classification. MPRI inherits the merits of the principle of relevant information (PRI) to effectively extract multiscale information embedded in the given data, and also takes advantage of the multilayer structure to learn representations in a coarse-to-fine manner. Specifically, MPRI performs spectral-spatial pixel characterization (using PRI) and  feature dimensionality reduction (using regularized linear discriminant analysis) iteratively and successively. Extensive experiments on three benchmark data sets demonstrate that MPRI outperforms existing state-of-the-art methods (including deep learning based ones) qualitatively and quantitatively, especially in the scenario of limited training samples. Code of MPRI is available at \url{http://bit.ly/MPRI_HSI}.
\end{abstract}

\section{Introduction}

With the rapid development of hyperspectral imaging techniques, current sensors always have high spectral and spatial resolution \cite{he2018recent}. For example, the ROSIS sensor can cover spectral resolution higher than $10$nm, reaching $1$m per pixel spatial resolution~\cite{cao2018hyperspectral,Zare2016Discriminative}. The increased spectral and spatial resolution enables us to accurately discriminate diverse materials of interest. As a result, hyperspectral images (HSIs) have been widely used in many practical applications, such as precision agriculture, environmental management, mining and mineralogy~\cite{he2018recent}. Among them, HSI classification, which aims to assign each pixel of HSI to a unique class label, has attracted increasing attention in recent years. However, the unfortunate combination of high-dimensional spectral features and the limited ground truth samples, as well as different atmospheric scattering conditions, make the HSI data inherently highly nonlinear and difficult to be categorized \cite{ghamisi2017advances}.

Early HSI classification methods straightforwardly apply conventional dimensionality reduction techniques, such as the principal component analysis (PCA) and the linear discriminant analysis (LDA), on spectral domain to learn discriminative spectral features. Although these methods are conceptually simple and easy to implement, they neglect the spatial information, a complement to spectral behavior that has been demonstrated effective to augment HSI classification performance~\cite{he2018recent,ghamisi2015survey}. To address this limitation, Chen {\it et al}.~\cite{chen2011hyperspectral} proposed the joint sparse representation (JSR) to incorporate spatial neighborhood information of pixels. Soltani-Farani {\it et al}.~\cite{A.Soltani-Farani15} designed spatial aware dictionary learning (SADL) by using a structured dictionary learning model to incorporate both spectral and spatial information. Kang {\it et al}. suggested using an edge-preserving filter (EPF) to improve the spatial structure of HSI~\cite{X.Kang14} and also introduced PCA to encourage the separability of new representations~\cite{kang2017pca}. A similar idea appears in Pan {\it et al.}~\cite{pan2017hierarchical}, in which EPF is substituted with a hierarchical guidance filter.
Although these methods perform well, the discriminative power of their extracted spectral-spatial features is far from satisfactory when being tested on challenging land covers.

A recent trend is to use deep neural networks (DNN), such as autoencoders (AE)~\cite{ma2016spectral} and convolutional neural networks (CNN)~\cite{Chen2016Deep}, to learn discriminative spectral-spatial features \cite{zhong2018spectral}. Although deep features always demonstrate superior discriminative power than hand-crafted features in different computer vision or image processing tasks, existing DNN based HSI classification methods either improve the performance marginally or require significantly more labeled data~\cite{yang2018hyperspectral}. On the other hand, collecting labeled data is always difficult and expensive in remote sensing community~\cite{Zare2016Discriminative}. Admittedly, transfer learning has the potential to alleviate the problem of limited labeled data, it still remains an open problem to construct a reliable relevance between the target domain and the source domain due to the large variations between HSIs obtained by different sensors with unmatched imaging bands and resolutions~\cite{zhu2017deep}.


Different from previous work, this paper presents a novel architecture, termed multiscale principle of relevant information (MPRI), to learn discriminative spectral-spatial features for HSI classification. MPRI inherits the merits of the principle of relevant information (PRI)~\cite[Chapter~8]{principe2010information}~\cite[Chapter~3]{rao2008unsupervised} to effectively extract multiscale information from given data, and also takes advantage of the multilayer structure to learn representations in a coarse-to-fine manner. To summarize, the major contributions of this work are threefold.


\begin{itemize}
\item {We demonstrate the capability of PRI, originated from the information theoretic learning (ITL)~\cite{principe2010information}, to characterize 3D pictorial structures in HSI data.}

\item {We generalize PRI into a multilayer structure to extract hierarchical representations for HSI classification. A multiscale scheme is also incorporated to model both local and global structures.}

\item {MPRI outperforms state-of-the-art HSI classification methods based on classical machine learning models (e.g., PCA-EPF~\cite{kang2017pca} and HIFI~\cite{pan2017hierarchical}) by a large margin. Using significantly fewer labeled data, MPRI also achieves almost the same classification accuracy compared to existing deep learning techniques (e.g., SAE-LR~\cite{Y.Chen14} and 3D-CNN \cite{li2017spectral}).}
\end{itemize}

The remainder of this paper is organized as follows. Section~\ref{chapter_PRI} reviews the basic objective of PRI and formulates PRI under the ITL framework. The architecture and optimization of our proposed MPRI is elaborated in Section~\ref{chapter_MPRI}. Section~\ref{chapter_experiment} shows experimental results on three popular HSI data sets. Finally, Section~\ref{chapter_conclusion} draws the conclusion.

\section{Elements of Renyi's $\alpha$-entropy and the principle of relevant information} \label{chapter_PRI}

Before presenting our method, we start with a brief review of the general idea and the objective of PRI, and then formulate this objective under the ITL framework.

\subsection{PRI: the general idea and its objective}

Suppose we are given a random variable $\bf{X}$ with a known probability density function (PDF) $g$, from which we want to learn a reduced statistical representation characterized by a random variable $\bf Y$ with PDF $f$. The PRI~\cite[Chapter~8]{principe2010information}~\cite[Chapter~3]{rao2008unsupervised} casts this problem as a trade-off between the entropy $H(f)$ of $\bf Y$ and its descriptive power about $\bf X$ in terms of their divergence $D(f\|g)$. Therefore, for a fixed PDF $g$, the objective of PRI is given by:
\begin{equation}
\underset{f}{\mathrm{minimize}}~H(f)+\beta D(f\|g),
\end{equation}where $\beta$ is a hyper-parameter controlling the amount of relevant information that $\bf Y$ can extract from $\bf X$. Note that, the minimization of entropy can be viewed as a means of finding the statistical regularities in the outcomes of a process, whereas the minimization of information theoretic divergence, such as the Kullback-Leibler divergence~\cite{kullback1951information} or the Chernoff divergence~\cite{chernoff1952measure}, ensuring that the regularities are closely related to $\bf X$. The PRI is similar in spirit to the Information Bottleneck (IB) method \cite{tishby2000information}, but the formulation is different because PRI does not require an observed relevant (or auxillary) variable and the optimization is done directly on the random variable $\bf X$, which provides a set of solutions that are related to the principal curves~\cite{hastie1989principal} of $g$, as will be demonstrated below.

\subsection{Formulation of PRI using Renyi's entropy functional}
In information theory, a natural extension of the well-known Shannon's entropy is the Renyi's $\alpha$-entropy \cite{renyi1961measures}. For a random variable $\bf{X}$ with PDF $f(x)$ in a finite set $\mathcal{X}$, the $\alpha$-entropy of $H(\bf{X})$  is defined as:

\begin{equation}
H_\alpha(f)=\frac{1}{1-\alpha}\log \int_\mathcal{X} f^\alpha(x)dx.
\label{1.1}
\end{equation}

On the other hand, motivated by the famed Cauchy-Schwarz (CS) inequality:

\begin{equation}
\Big| \int f(x)g(x)dx \Big|^2 \leq \int \mid f(x)\mid^2 dx \int \mid g(x)\mid^2 dx,
\end{equation}
with equality if and only if $f(x)$ and $g(x)$ are linearly dependent (e.g., $f(x)$ is just a scaled version of $g(x)$), a measure of the ``distance'' between the PDFs can be defined, which was named the CS divergence~\cite{jenssen2006cauchy}, with:

\begin{equation} \label{1.2}
\begin{split}
D_{cs} (f\|g) & = -\log(\int fg)^2 + \log(\int f^2) + \log(\int g^2) \\
& = 2H_2(f;g) - H_2(f) - H_2(g),
\end{split}
\end{equation}
the term $H_2(f;g)=-\log\int f(x)g(x)dx$ is also called the quadratic cross entropy~\cite{principe2010information}.

Combining Eqs.~(\ref{1.1}) and (\ref{1.2}), the PRI under the $2$-order Renyi's entropy can be formulated as:

\begin{equation}\label{1.3}
\begin{aligned}
f_{\text{opt}}& =\arg \min_f H_2(f)+\beta(2H_2(f;g)-H_2(f)-H_2(g))\\
& \equiv \arg \min_f (1-\beta)H_2(f) + 2\beta H_2(f;g),
\end{aligned}
\end{equation}
the second equation holds because the extra term $\beta H_2(g)$ is a constant with respect to $f$. 


Given $\mathbf{X}=\{\mathbf{x}_i\}_{i=1}^N$ and $\mathbf{Y}=\{\mathbf{y}_i\}_{i=1}^N$, both in $\mathbb{R}^p$, drawn~\emph{i.i.d.} from $g$ and $f$, respectively. Using the Parzen-window density estimation~\cite{parzen1962estimation} with Gaussian kernel $G_{\delta}(\cdot)=\exp(-\frac{\|\cdot\|^2}{2\delta^2})$, Eq.~(\ref{1.3}) can be simplified as~\cite{rao2008unsupervised}:

\begin{equation}\label{PRI_formulation}
\begin{aligned}
{\bf Y}_{\text{opt}}&=&\arg \min_{\bf Y}[-(1-\beta)\log\big(\frac{1}{N^2}\sum_{i,j=1}^NG_{\delta}({\bf y}_i-{\bf y}_j)\big)\\
&&-2\beta \log\big(\frac{1}{N^2}\sum_{i,j=1}^NG_{\delta}({\bf y}_i-{\bf x}_j)\big)].
\end{aligned}
\end{equation}

\begin{figure}[h]
\centering \subfigure[original data]{
\includegraphics[height = 0.9in]{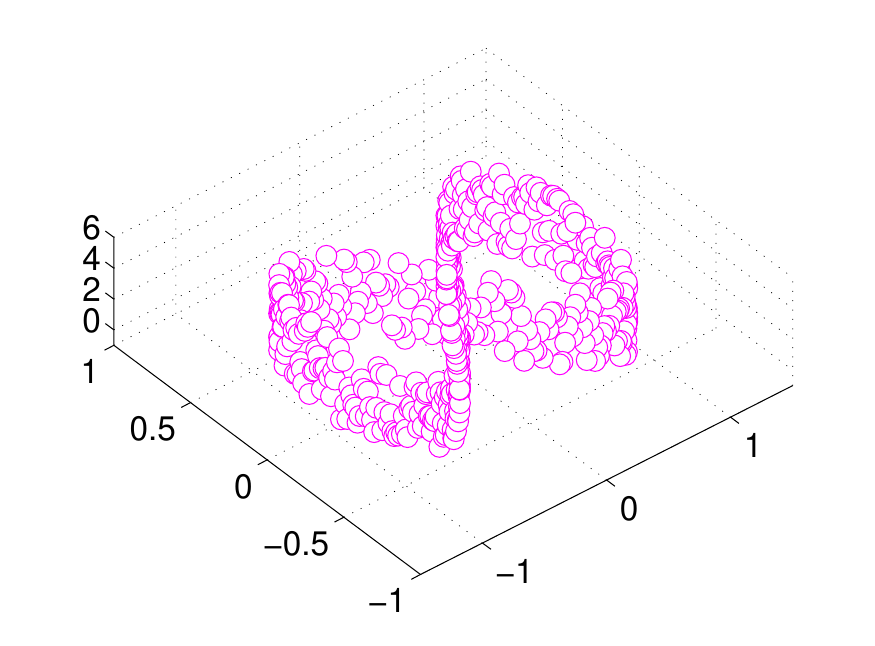}}
\centering \subfigure[$\beta=0$]{
\includegraphics[height =0.9in]{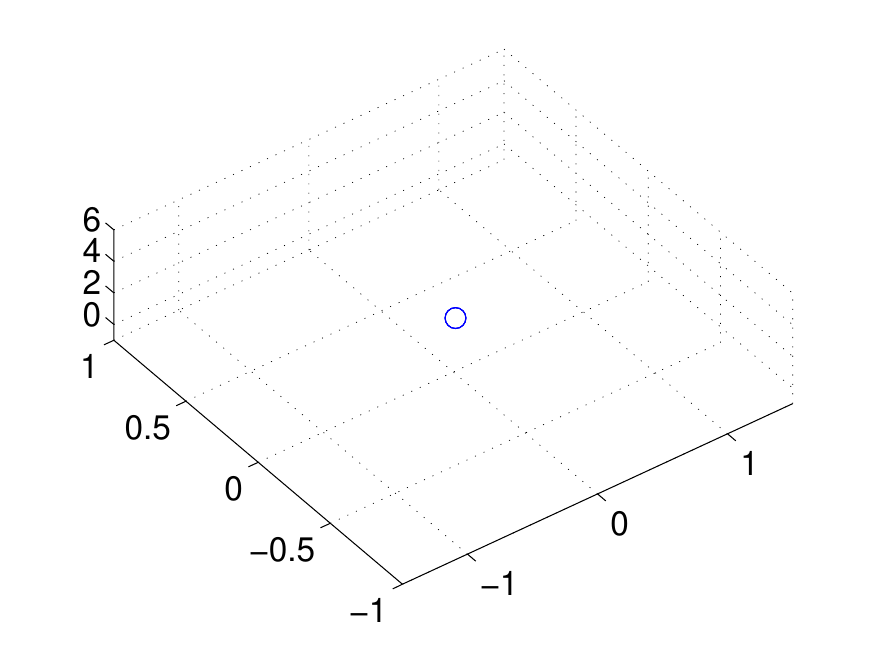}}
\centering \subfigure[$\beta=1$]{
\includegraphics[height =0.9in]{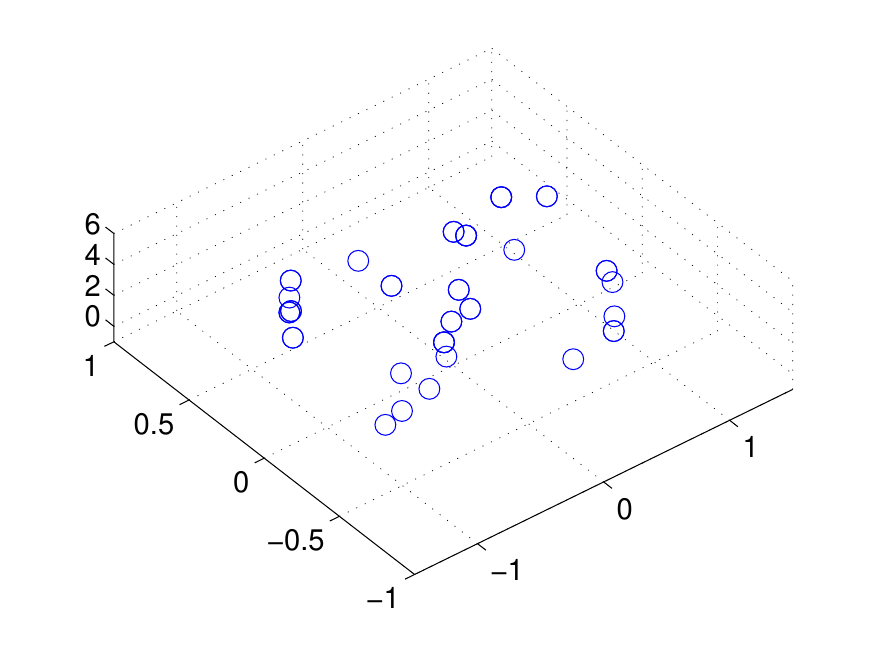}}\\
\centering \subfigure[$\beta=3$]{
\includegraphics[height =0.9in]{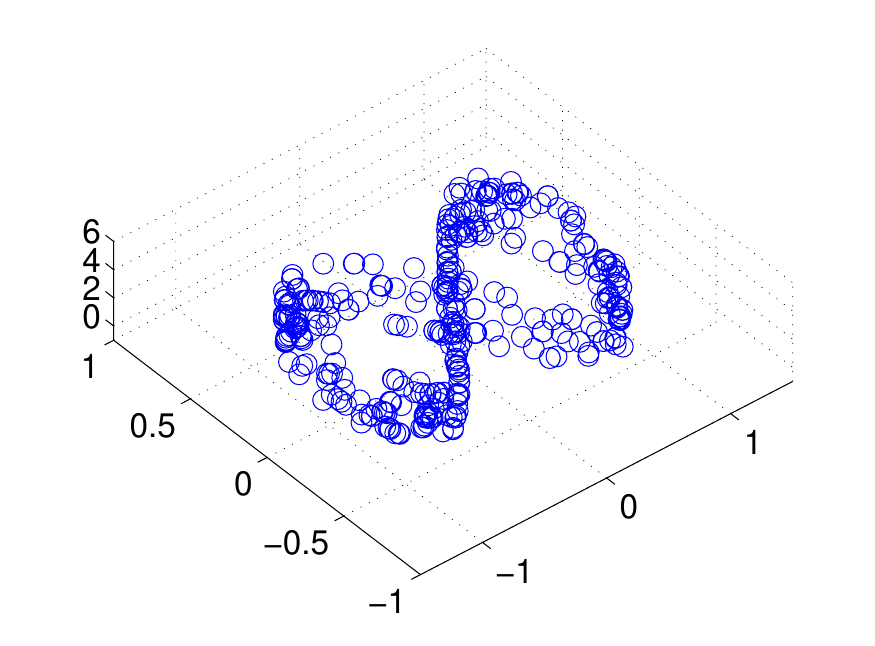}}
\centering \subfigure[$\beta=6$]{
\includegraphics[height = 0.9in]{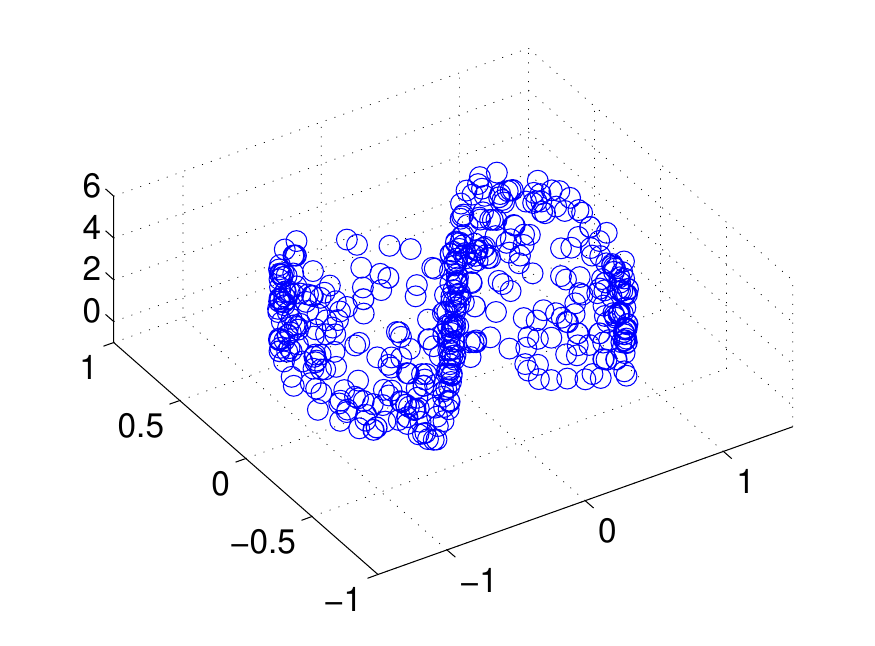}}
\centering \subfigure[$\beta=100$]{
\includegraphics[height =0.9in]{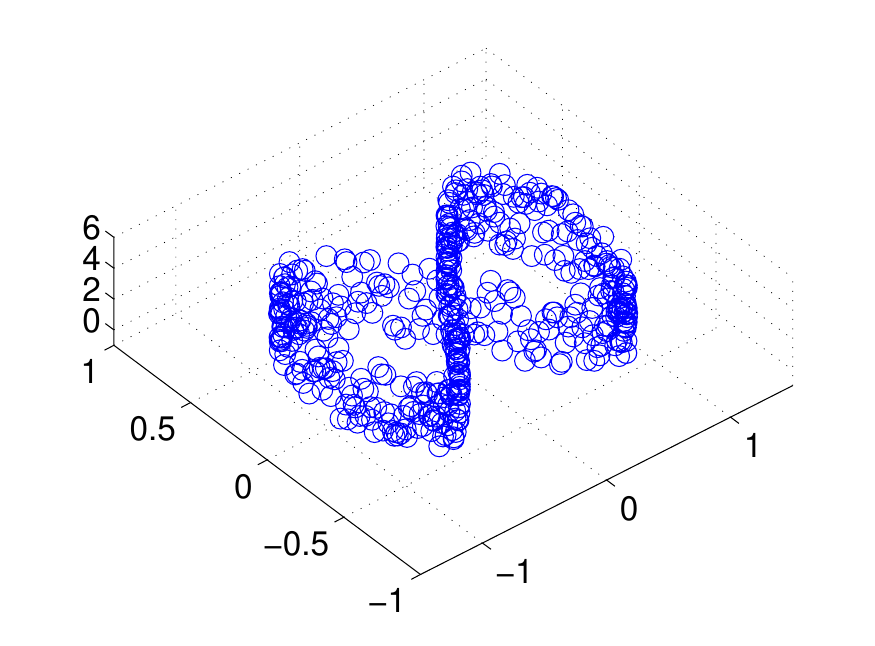}}
\caption{Illustration of the structures revealed by the PRI for (a) Intersect data set. As the values of $\beta$ increase the solution passes through (b) a single point, (c) modes, (d) and (e) principal curves at different dimensions, and in the extreme case of (f) $\beta \rightarrow \infty$ we get back the data themselves as the solution.} \label{EX}
\end{figure}

It turns out that the value of $\beta$ defines various levels of information reduction, ranging from data mean value ($\beta =0$), clustering ($\beta =1$), principal curves \cite{hastie1989principal} extraction at different dimensions, and vector quantization obtaining back the initial data when $\beta \rightarrow \infty$~\cite{principe2010information,rao2008unsupervised}. Hence, the PRI achieves similar effects to a moment decomposition of the PDF controlled by a single parameter $\beta$, using a data driven optimization approach.
See Fig.~\ref{EX} for an example.
From this figure we can see that the self organizing decomposition provides a set of hierarchical features of the input data beyond cluster centers, that may yield more robust features.
Note that, despite its strategic flexibility to find reduced structure of given data, the PRI is mostly unknown to practitioners.

\section{Multiscale principle of relevant information (MPRI) for HSI classification} \label{chapter_MPRI}

\begin{figure}[htb]
\centering
\includegraphics[width=4.8in]{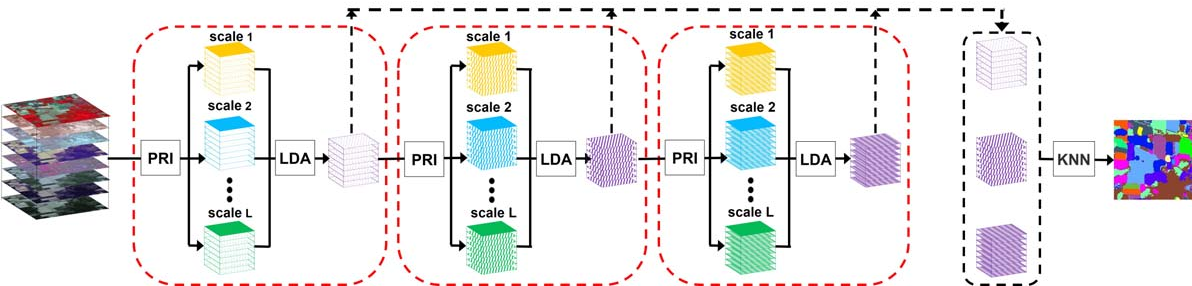}
\caption{The architecture of multiscale principle of relevant information (MPRI) for HSI classification. The spectral-spatial feature learning unit is marked with red dashed rectangle. The spectral-spatial features are extracted by performing PRI (in multiple scales) and LDA iteratively and successively on HSI data cube (after normalization). Finally, features from each unit are concatenated and fed into a $k$-nearest neighbors (KNN) classifier to predict pixel labels. This plot only demonstrates a $3$-layer MPRI, but the number of layers can be increased or decreased flexibly.}\label{MPRI}
\end{figure}

In this section, we present MPRI for HSI classification. MPRI stacks multiple spectral-spatial feature learning units, in which each unit consists of multiscale PRI and a regularized LDA~\cite{bandos2009classification}. The architecture of MPRI is shown in Fig.~\ref{MPRI}.


To the best of our knowledge, apart from performing band selection (e.g.,~\cite{feng2015mutual,yu2019multivariate}) or measuring spectral variability (e.g.,~\cite{chang2000information}), information theoretic principles have seldom been investigated to learn discriminative spectral-spatial features for HSI classification. The most similar work to ours is~\cite{kamandar2013linear}, in which the authors use the criterion of minimum redundancy maximum relevance (MRMR)~\cite{peng2005feature} to extract linear features. However, owing to the poor approximation to estimate multivariate mutual information, the performance of~\cite{kamandar2013linear} is only slightly better than the basic linear discriminant analysis (LDA)~\cite{du2007modified}.

\subsection{Spectral-Spatial Feature Learning Unit}\label{learning_unit}

\begin{figure}[htbp]
\centering
\includegraphics[width = 4.5in]{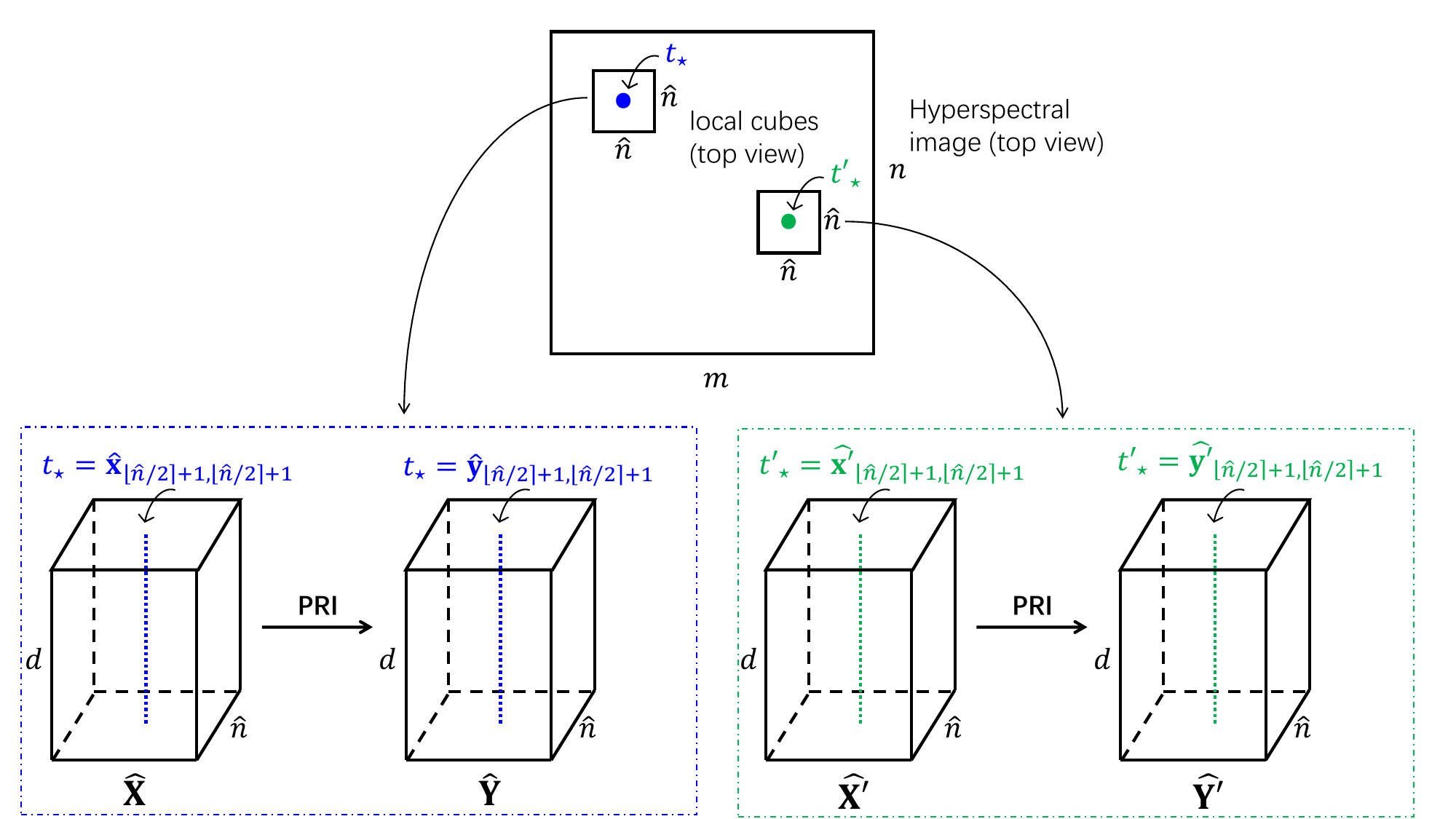}
\caption{For each target spectral vector (e.g., $\bf{t}_\star$ or $\bf{t}'_\star$) in the raw hyperspectral image, we obtain a new vector representation by performing PRI in its corresponding local data cube (e.g., $\hat{\bf X}$ or $\hat{\bf X}'$).} \label{sliding}
\end{figure}

Let $\mathbf{T}\in \mathbb{R}^{m\times n \times d}$ be the raw 3D HSI data cube, where $m$ and $n$ are the spatial dimensions, $d$ is the number of spectral bands. For a target spectral vector $\mathbf{t}_\star\in \mathbb{R}^d$, we extract a local cube (denote $\hat{\bf X}$) from $\bf{T}$ using a sliding window of width $\hat{n}$ centered at $\bf{t}_\star$, i.e., $\hat{\bf X}=\{\hat{\bf x}_1, \hat{\bf x}_2, \cdots, \hat{\bf x}_{\hat{N}}\}\in \mathbb{R}^{\hat{N}\times d}$, $\hat{n}\times\hat{n}=\hat{N}$, and $\bf{t}_\star=\hat{\bf x}_{\nint{\hat{n}/2}+1,\nint{\hat{n}/2}+1}$, where $\nint{\cdot}$ is the nearest integer function.
We obtain the spectral-spatial characterization $\hat{\bf Y}=\{\hat{\bf y}_1, \hat{\bf y}_2, \cdots, \hat{\bf y}_{\hat{N}}\}\in \mathbb{R}^{\hat{N}\times d}$ from $\hat{\bf X}$ using PRI via the following objective:
\begin{equation}\label{eq:PRI}
\begin{aligned}
{\rm minimize}_{\bf \hat{Y}}[-(1-\beta)\log\big(\frac{1}{\hat{N}^2}\sum_{i,j=1}^{\hat{N}}G_{\delta}(\hat{\bf y}_i-\hat{\bf y}_j)\big) -2\beta \log\big(\frac{1}{\hat{N}^2}\sum_{i,j=1}^{\hat{N}}G_{\delta}(\hat{\bf y}_i-\hat{\bf x}_j)\big)].
\end{aligned}
\end{equation}
We finally use the center vector of $\hat{\bf Y}$, i.e., $\hat{\bf y}_{\nint{\hat{n}/2}+1,\nint{\hat{n}/2}+1}$, as the new representation of $\bf{t}_\star$. We scan the whole 3D cube with a sliding window of width $\hat{n}$ targeted at each pixel to get the new spectral-spatial representation. The procedure is depicted in Fig.~\ref{sliding}.

Eq.~(\ref{eq:PRI}) is updated iteratively. Specifically,
denote $V(\hat{\bf Y})=\frac{1}{{\hat N}^2}\sum_{i,j=1}^{\hat N}G_\delta(\hat{\bf y}_i-\hat{\bf y}_j)$
and
$V(\hat{\bf Y}; \hat{\bf X})=\frac{1}{\hat N^2}\sum_{i,j=1}^{\hat N}G_\delta(\hat{\bf y}_j-\hat{\bf x}_i)$,
taking the derivative of Eq.~(\ref{eq:PRI}) with respect to $\hat{\bf y}_\star$ and equating to zero, we have:
\begin{equation}
\begin{aligned}
\frac{1-\beta}{V(\hat{\bf Y})} \sum_{j=1}^{\hat N} G_\delta(\hat{\bf y}_\star-\hat{\bf y}_j)\left\{\frac{\hat{\bf y}_j-\hat{\bf y}_\star}{\delta^2}\right\}
+ \frac{\beta}{V(\hat{\bf Y};\hat{\bf X})} \sum_{j=1}^{\hat N} G_\delta(\hat{\bf y}_\star-\hat{\bf x}_j)\left\{ \frac{\hat{\bf x}_j-\hat{\bf y}_\star}{\delta^2}\right\}=0.
\label{xkk}
\end{aligned}
\end{equation}

Rearrange Eq.~(\ref{xkk}), we have:
\begin{equation}
\begin{aligned}
\{\frac{\beta}{V(\hat{\bf Y};\hat{\bf X})}\sum_{j=1}^{\hat N} G_\delta(\hat{\bf y}_\star-\hat{\bf x}_j) \}\hat{\bf y}_\star
=\frac{1-\beta}{V(\hat{\bf Y})} \sum_{j=1}^{\hat N} G_\delta(\hat{\bf y}_\star-\hat{\bf y}_j){\hat{\bf y}_j}\\
\quad\quad\quad\quad\quad-\{\frac{1-\beta}{V(\hat{\bf Y})}\sum_{j=1}^{\hat N} G_\delta(\hat{\bf y}_\star-\hat{\bf y}_j)\}{\hat{\bf y}_\star}
+\frac{\beta}{V(\hat{\bf Y};\hat{\bf X})}\sum_{j=1}^{\hat N} G_\delta(\hat{\bf y}_\star-\hat{\bf x}_j)\hat{\bf x}_j.
\label{xkk2}
\end{aligned}
\end{equation}

Divide both sides of the Eq.~(\ref{xkk2}) by
\begin{equation}
\frac{\beta}{V(\hat{\bf Y};\hat{\bf X})} \sum_{j=1}^{\hat N} G_\delta(\hat{\bf y}_\star-\hat{\bf x}_j),
\end{equation}
and let
\begin{equation}
c={V(\hat{\bf Y};\hat{\bf X})}/{V(\hat{\bf Y})},
\end{equation}
we obtain the fixed point update rule for $\hat{\bf y}_\star$:

\begin{equation}
\begin{aligned}
\hat{\bf y}_\star^{\tau+1}=c\frac{1-\beta}{\beta}\frac{\sum_{j=1}^{\hat N} G_{\delta}(\hat{\bf y}_\star^\tau-\hat{\bf y}_j^\tau)\hat{\bf y}_j^\tau}{\sum_{j=1}^{\hat N} G_{\delta}(\hat{\bf y}_\star^\tau-\hat{\bf x}_j)}
-c\frac{1-\beta}{\beta}\frac{\sum_{j=1}^{\hat N} G_{\delta}(\hat{\bf y}_\star^\tau-\hat{\bf y}_j^\tau)}{\sum_{j=1}^{\hat N} G_{\delta}(\hat{\bf y}_\star^\tau-\hat{\bf x}_j)}\hat{\bf y}_\star^\tau\\
+\frac{\sum_{j=1}^{\hat N} G_{\delta}(\hat{\bf y}_\star^\tau-\hat{\bf x}_j)\hat{\bf x}_j}{\sum_{j=1}^{\hat N} G_{\delta}(\hat{\bf y}_\star^\tau-\hat{\bf x}_j)},
\label{xk}
\end{aligned}
\end{equation}
where $\tau$ is the iteration number. We move the sliding window pixel by pixel, and only update the representation of the center target pixel, as shown in Fig.~\ref{sliding}.

We also introduce two modifications to increase the discriminative power of the new representation. First, different values of $\hat{n}$ ($3$, $5$, $7$, $9$, $11$, $13$ in this work) are used to model both local and global structures. Second, to reduce the redundancy of raw features constructed by concatenating PRI representations in multiple scales, we further perform a regularized LDA~\cite{bandos2009classification}.

Note that, the hyper-parameter $\beta$ and different values of $\hat{n}$ play different roles in MPRI.
Specifically, $\beta$ in PRI balances the trade-off between the regularity of extracted representation and its discriminative power to the given data. Therefore, it should be set in a reasonable range to avoid over-smoothing effect of the resulting image and unsatisfactory classification performance. A deeper discussion is shown in Section~\ref{sec:beta_effect}.
By contrast, $\hat{n}$ controls the spatial scale of the learned representation. The motivation is that the discriminative information of different categories may not be easily characterized by a sliding window of a fixed size (i.e., $\hat{n}\times \hat{n}$). Thus, it would be favorable if one can incorporate discriminative information from different scales, by changing the value of $\hat{n}$.




\subsection{Stacking Multiple Units}



In order to characterize spectral-spatial structures in a coarse-to-fine manner, we stack multiple spectral-spatial feature learning units described in Section~\ref{learning_unit} to constitute a multilayer structure and concatenate representations from each layer to form the final spectral-spatial representation. We finally feed this representation into a standard $k$-nearest neighbors (KNN) for classification.

Different from existing DNNs that are typically trained with error backpropagation or the combination of a greedy layer-wise pretraining and a fine-tuning stage, our multilayer structure is trained successively from bottom layer to top layer without error backpropagation. For the $i$-th layer, the input of PRI is the representation learned from the previous layer (denoted $T_{i-1}$). We then learn new representation $T_i$ by iteratively updates $T_{i-1}$ with Eq.~(\ref{xk}) and a dimensionality reduction step with LDA at the end of iteration. As for the multiscale PRI, it can be trained in parallel with respect to different sliding window sizes.

\begin{figure}[!htbp]
\centering \subfigure[]{
\includegraphics[width = 1.5in]{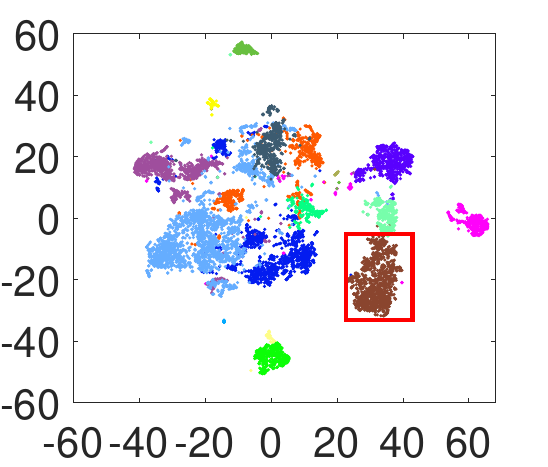}}
\centering \subfigure[]{
\includegraphics[width = 1.5in]{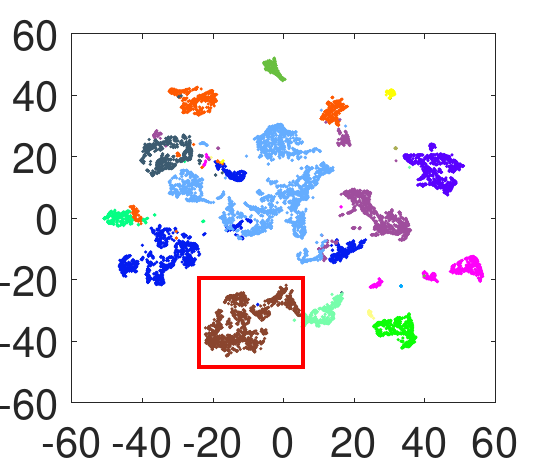}}
\subfigure[]{
\includegraphics[width = 1.5in]{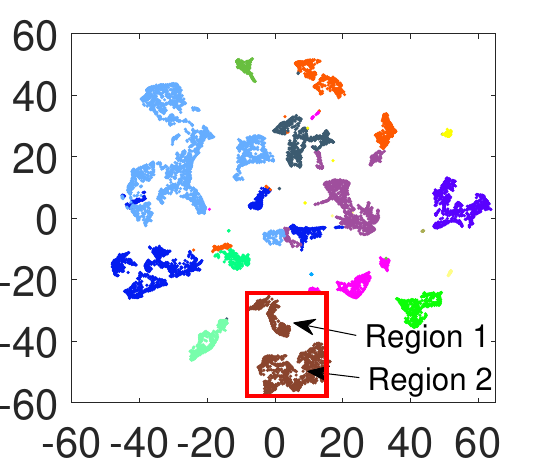}} \\
\centering \subfigure[]{
\includegraphics[width = 1.5in]{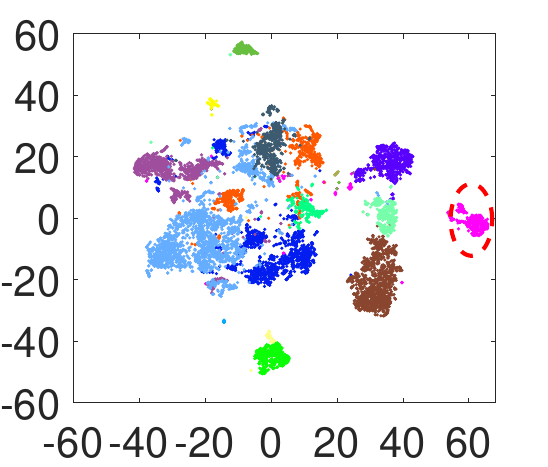}}
\centering \subfigure[]{
\includegraphics[width = 1.5in]{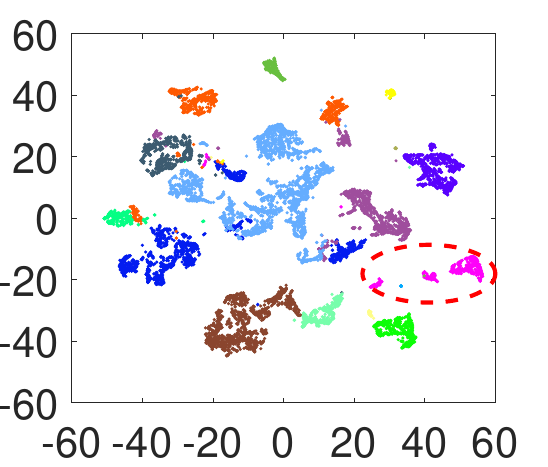}}
\subfigure[]{
\includegraphics[width = 1.5in]{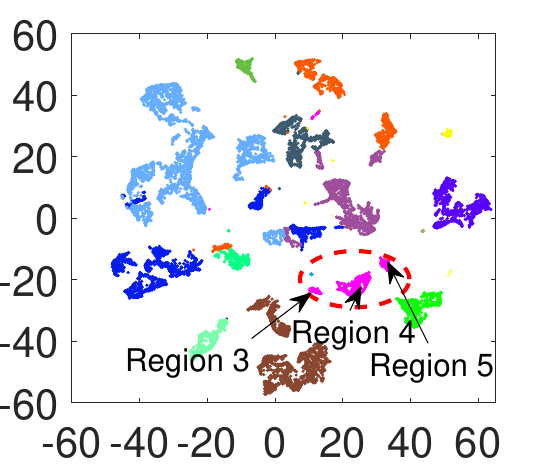}} \\
\centering \subfigure[]{
\includegraphics[height = 1.5in]{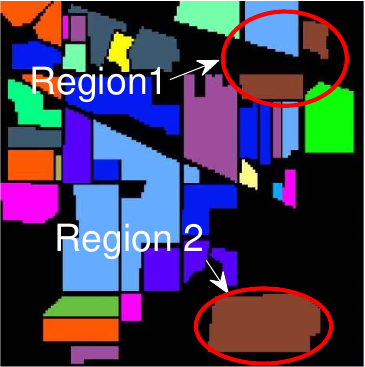}}
\centering \subfigure[]{
\includegraphics[height = 1.5in]{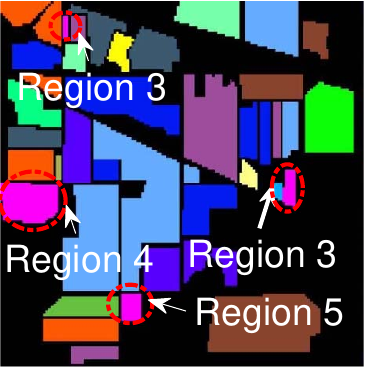}}
\hspace{10pt}\centering \subfigure[]{
\includegraphics[height = 1.5in]{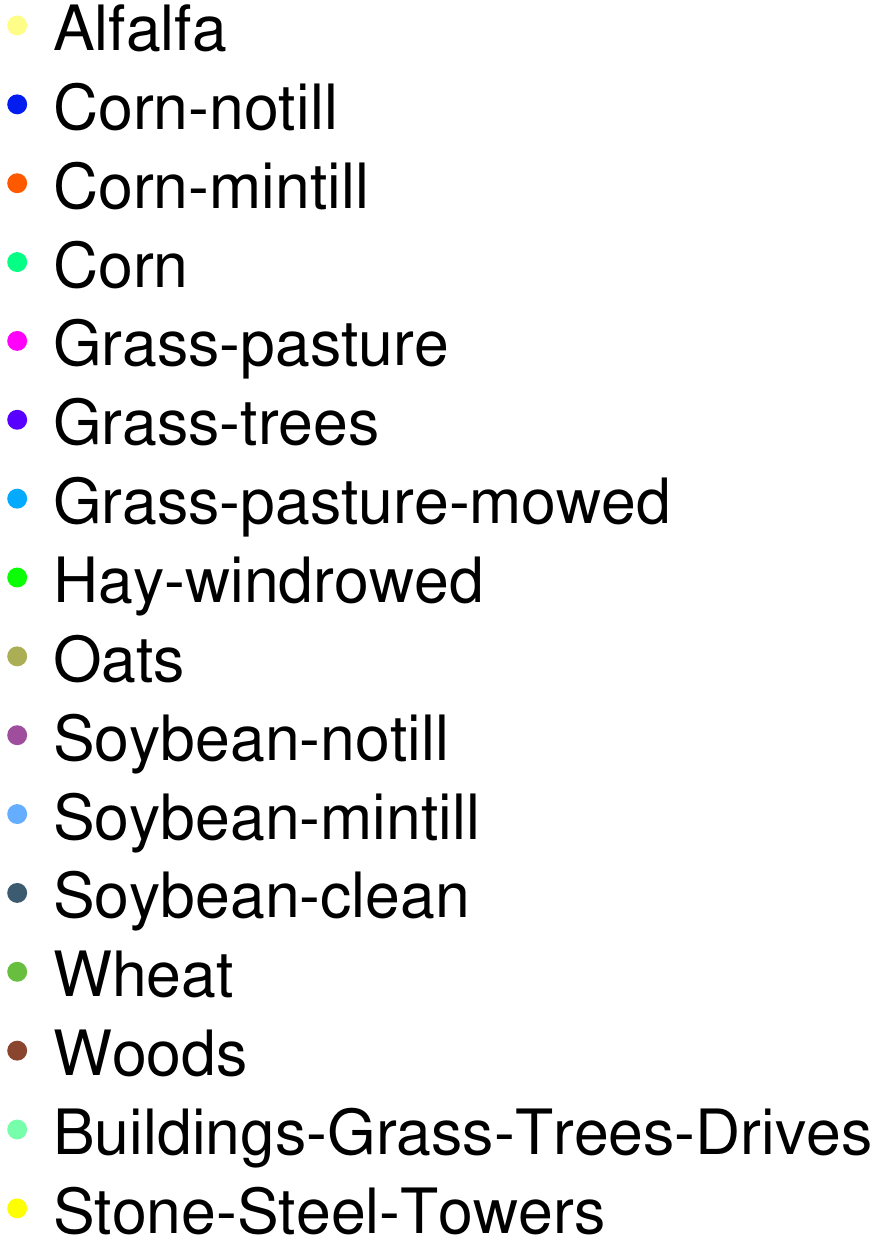}}\hspace{10pt}
\caption{2D projection of features learned by MPRI in different layers on Indian Pines data set. Features of ``Woods'' in the $1$st layer, the $2$nd layer, and the $3$rd layer are marked with red rectangle in (a)-(c). Similarly, features of ``Grass-pasture'' are marked with magenta ellipses in (d)-(f). (g) shows the locations of ``Region 1" and ``Region 2". (h) shows the locations of ``Region 3" , ``Region 4" and ``Region 5". (i) shows class legend.} \label{PF}
\end{figure}

The interpretation of DNN as a way of creating successively better representations of the data has already been suggested and explored by many (e.g.,~\cite{achille2018information}). Most recently, Schwartz-Ziv and Tishby~\cite{shwartz2017opening} put forth an interpretation of DNN as creating sufficient representations of the data that are increasingly minimal. For our deep architecture, in order to have an intuitive understanding to its inner mechanism, we plot the 2D projection (after $1,000$ t-SNE~\cite{maaten2008visualizing} iterations) of features learned from different layers in Fig.~\ref{PF}. Similar to DNN, MPRI creates successively more faithful and separable representations in deeper layers. Moreover, the deeper features can discriminate the with-in class samples in different geography regions, even though we do not manually incorporate geographic information in the training.


\section{Experimental Results} \label{chapter_experiment}


We conduct three groups of experiments to demonstrate the effectiveness and superiority of the MPRI. Specifically, we first perform a simple test to determine a reliable range for the value of $\beta$ in PRI and the number of layers in MPRI. Then, we implement MPRI and several of its degraded variants to analyze and evaluate component-wise contributions to performance gain. Finally, we evaluate MPRI against state-of-the-art methods on benchmark data sets using both visual and qualitative evaluations.


Three popular data sets, namely the Indian Pines~\cite{PURR1947}, the Pavia University and the Pavia Center, are selected in this work. We summarize the properties of each data set in Table~\ref{dataset}.
\begin{table}[!htbp]
\caption{Details of data sets.}
\centering
\setlength{\tabcolsep}{1mm}
{\begin{tabular}{l c c c}
\toprule
Data set &Indian Pines & Pavia University  & Pavia Center\\
\midrule
Sensor&AVIRIS&ROSIS&ROSIS-3\\
Spatial size&$145\times145$&610$\times$ 340&$1096\times 492$\\
$\sharp$ bands (used) &200&103&102\\
$\sharp$ classes&16&9&9\\
\bottomrule
\end{tabular}}
\label{dataset}
\end{table}

\begin{enumerate}
  \item  The first image, displayed in Fig. \ref{Indian Pines_data}(a), is called Indian Pines. It was gathered by the airborne visible/infrared imaging spectrometer (AVIRIS) sensor over the agricultural Indian Pines test site in northwestern Indiana,
United States. The size of this image is $145\times145$ pixels with spatial resolution of 20 m. The low spatial resolution leads to the presence of highly mixed pixels \cite{ghamisi2014automatic}. A three-band false color image and the ground-truth map are shown in Fig. \ref{Indian Pines_data}(a)-(b), where there are 16 classes of interest. And the name and quantity of each class are reported in Fig. \ref{Indian Pines_data}(c). The number of bands has been reduced to 200 by removing 20 bands covering the region of water absorption. This scene constitutes a challenging classification problem due to the significant presence of mixed pixels in all available classes and the unbalanced number of available labeled pixels per class \cite{J.Li13}.

\begin{figure}[!htbp]
\centering \subfigure[]{
\includegraphics[height = 1.25in]{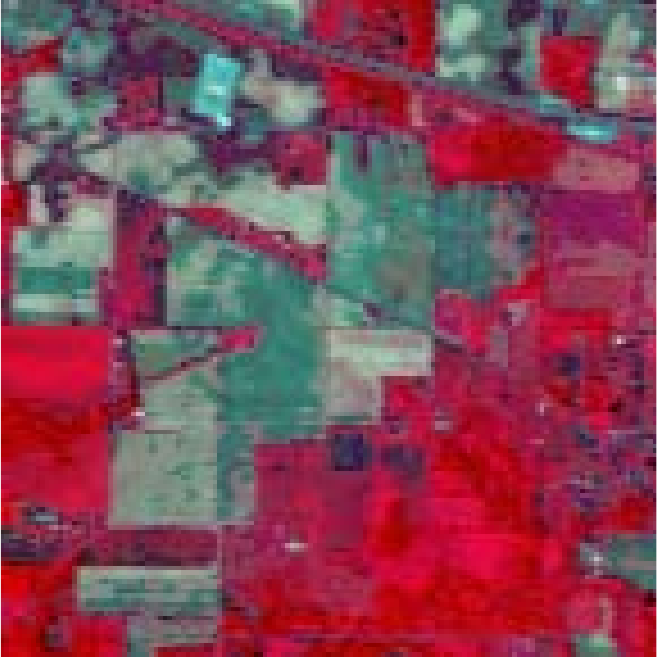}}
\centering \subfigure[]{
\includegraphics[height = 1.25in]{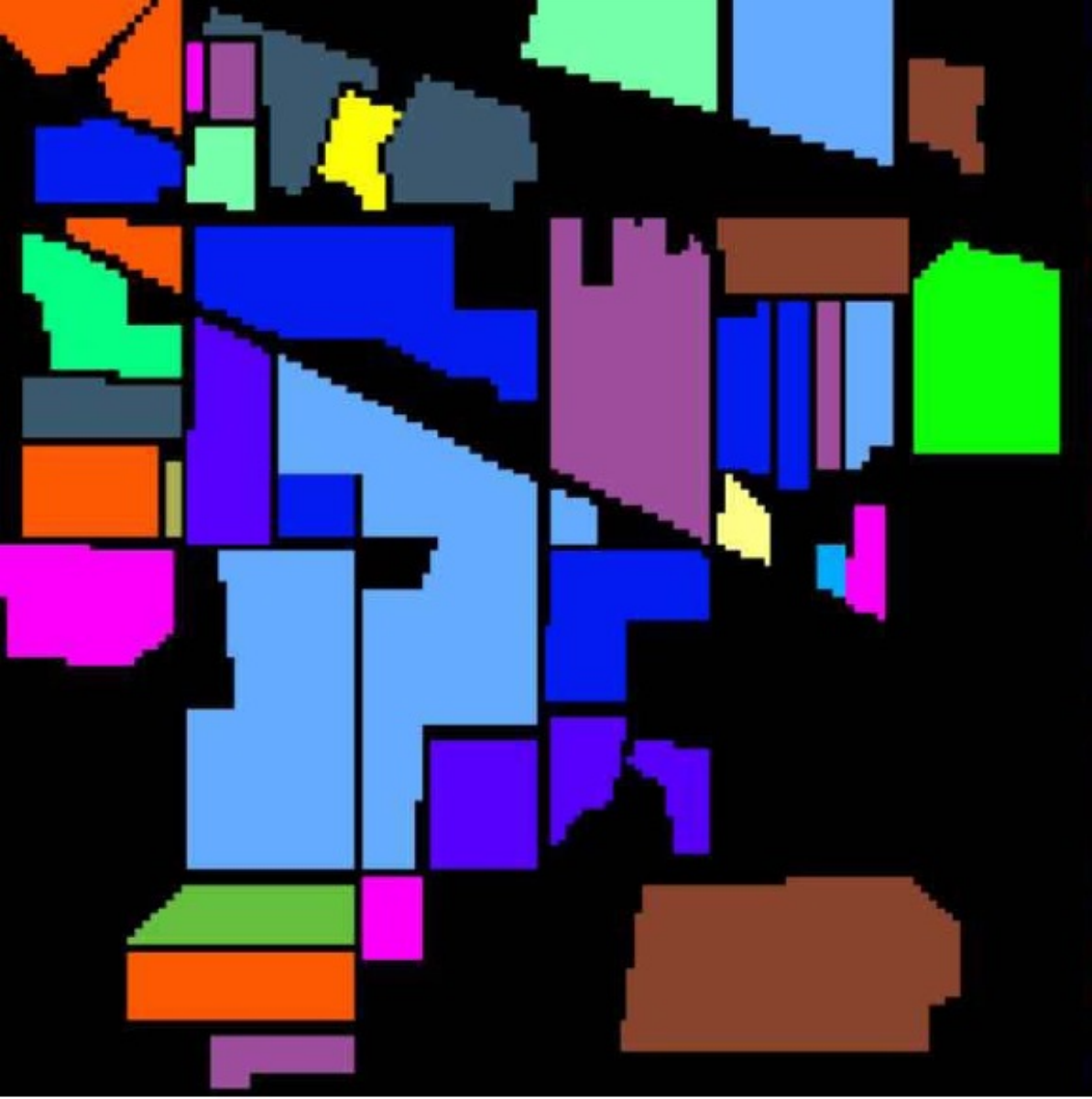}}
\centering \subfigure[]{
\includegraphics[width = 2in]{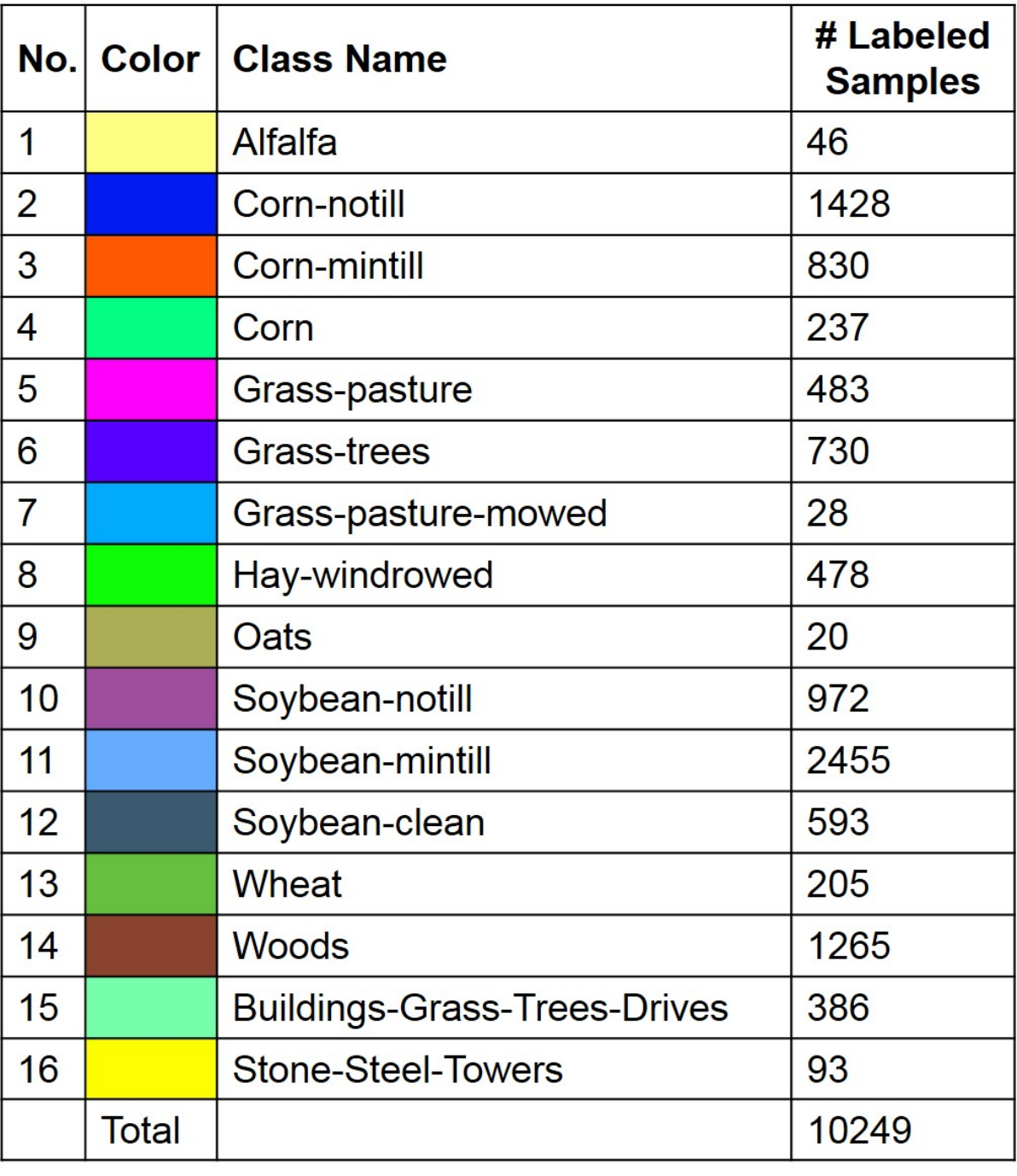}}
\caption{(a) False color composition of the AVIRIS Indian Pines scene. (b) Reference map containing 16 mutually exclusive land-cover classes. (c)The numbers of the labeled samples.} \label{Indian Pines_data}
\end{figure}

\item The second image is the Pavia University, which was recorded by the reflective optics spectrographic imaging system (ROSIS) sensor during a flight campaign over Pavia, northern Italy. This scene has 610$\times$ 340 pixels with a spatial resolution of 1.3m (covering the wavelength range from 0.4 to 0.9$\mu m$). There are 9 ground-truth classes, including trees, asphalt, bitumen, gravel, metal sheet, shadow, bricks, meadow, and soil.  In our experiments, 12 noisy bands have been removed and finally 103 out of the 115 bands were used. The class descriptions and sample distributions for this image are given in Fig. \ref{Pavia_U_data} (c). As can be seen, the total number of labeled samples in this image is 43923. A three-band false color image and the ground-truth map are also shown in Fig. \ref{Pavia_U_data}.

\begin{figure}[!htbp]
\centering \subfigure[]{
\includegraphics[width = 1.2in]{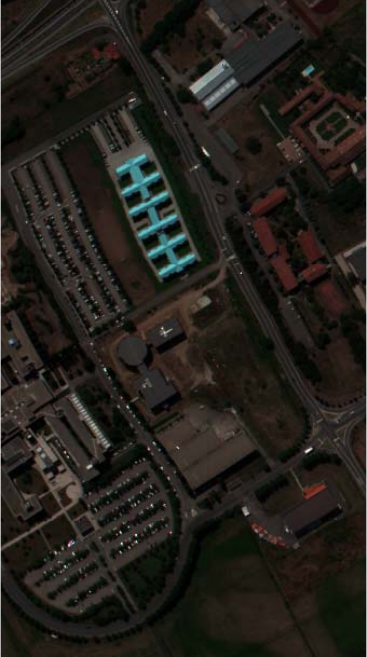}}
\centering \subfigure[]{
\includegraphics[width = 1.2in]{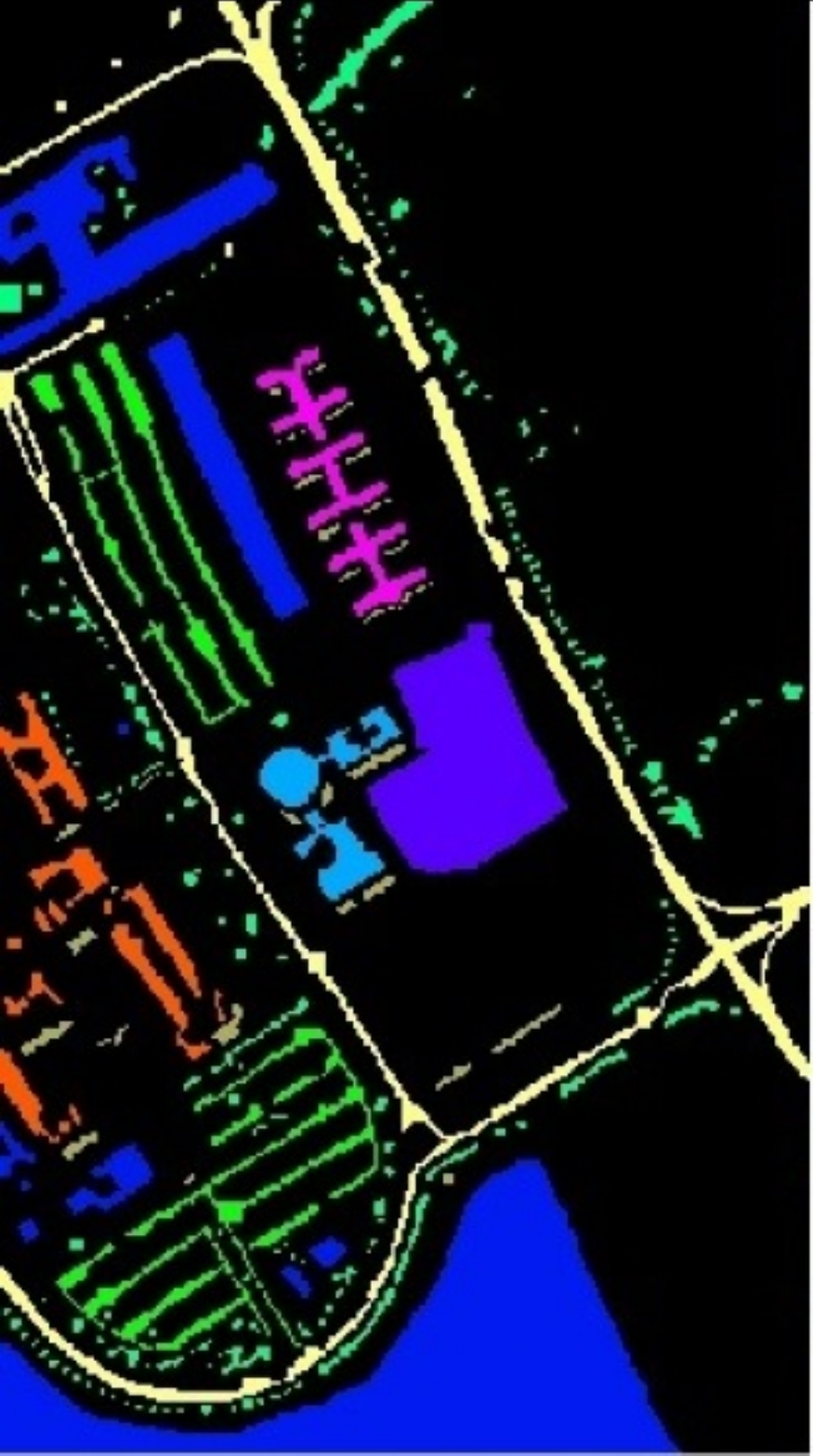}}
\centering \subfigure[]{
\includegraphics[width = 2.5in]{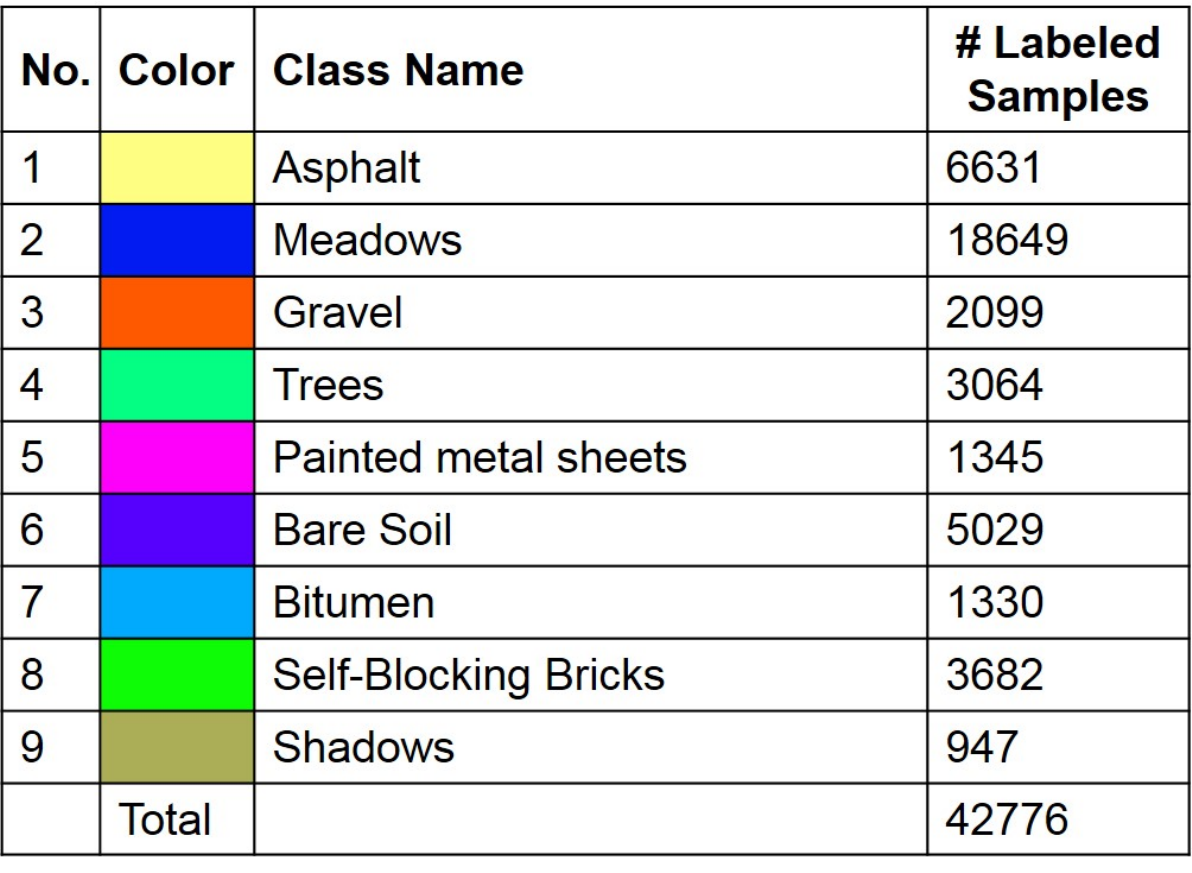}}
\caption{(a) False color composition of the Pavia University scene. (b) Reference map containing 9 mutually exclusive land-cover classes. (c) The numbers of the labeled samples.} \label{Pavia_U_data}
\end{figure}

\item The third data set is Pavia Center. It was acquired by ROSIS-3 sensor in 2003, with a spatial resolution of 1.3m and 102 spectral bands (some bands have been removed due to noise). A three-band false color image and the ground-truth map are also shown in Figs. \ref{Pavia_data} (a)- (b). The number
of ground truth classes is 9 (see Fig. \ref{Pavia_data}) and it consists of $1096\times 492$ pixels. The number of samples of each class ranges
from 2108 to 65278 (Fig. \ref{Pavia_data}(e)). There are 5536 training samples and 98015 testing samples (Fig. \ref{Pavia_data}(c)-(d)). Note that these training samples
are out of the testing samples.

\begin{figure}[!htbp]
\centering \subfigure[]{
\includegraphics[height = 2.6in]{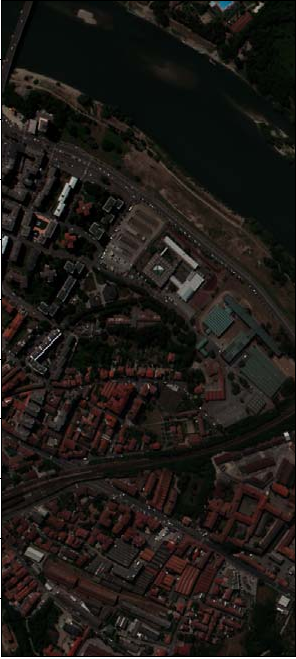}}
\centering \subfigure[]{
\includegraphics[height = 2.6in]{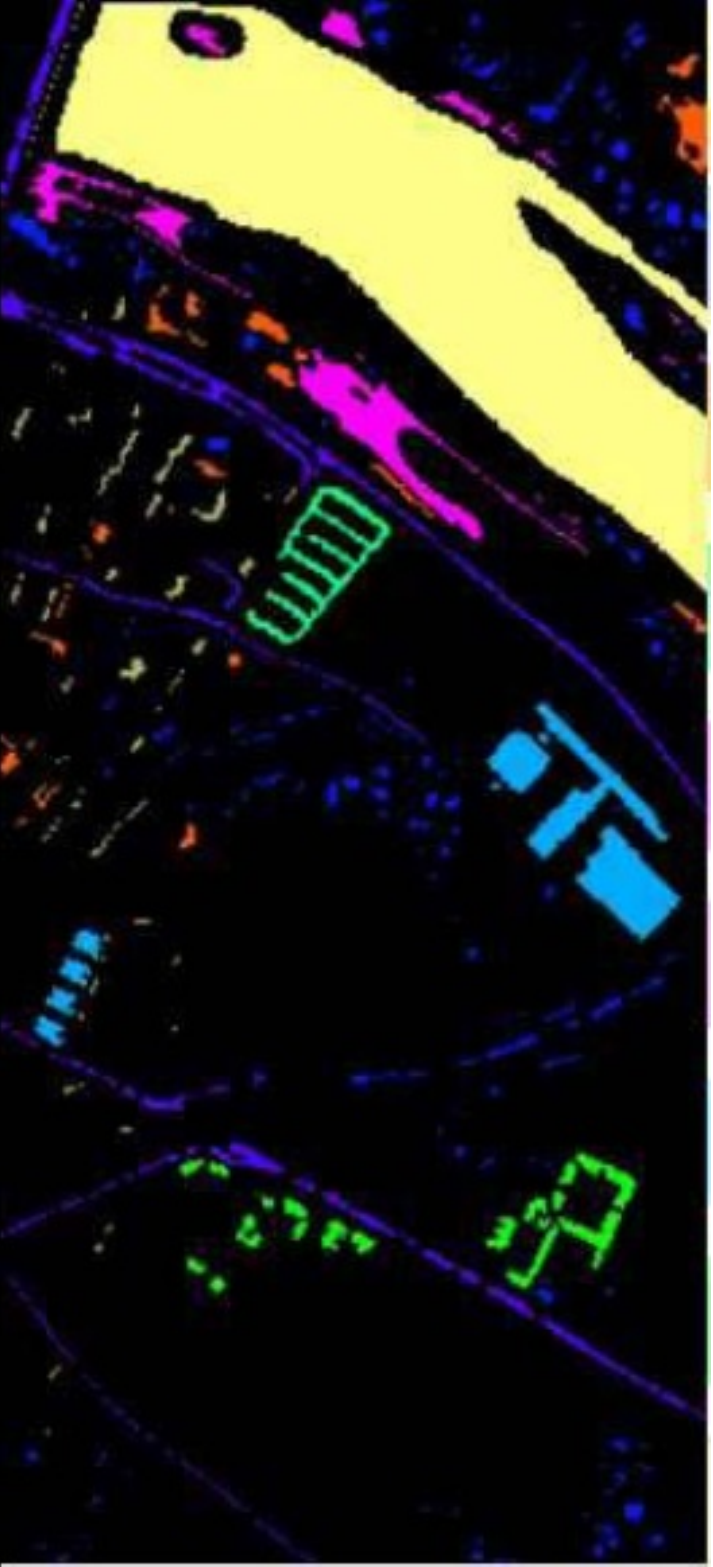}}
\centering \subfigure[]{
\includegraphics[height = 2.6in]{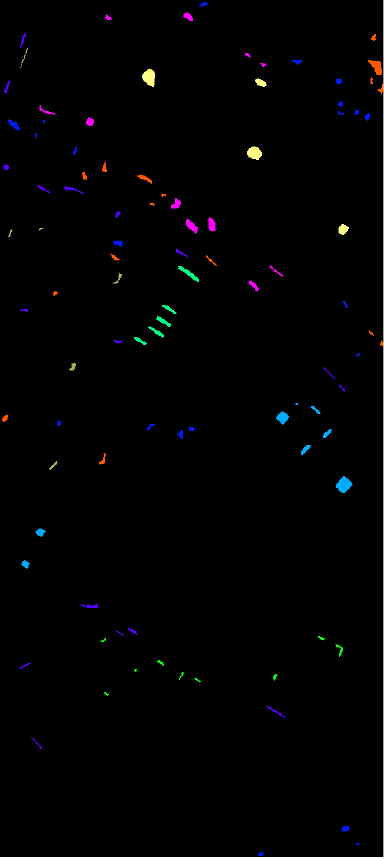}}\\
\centering \subfigure[]{
\includegraphics[height = 2.6in]{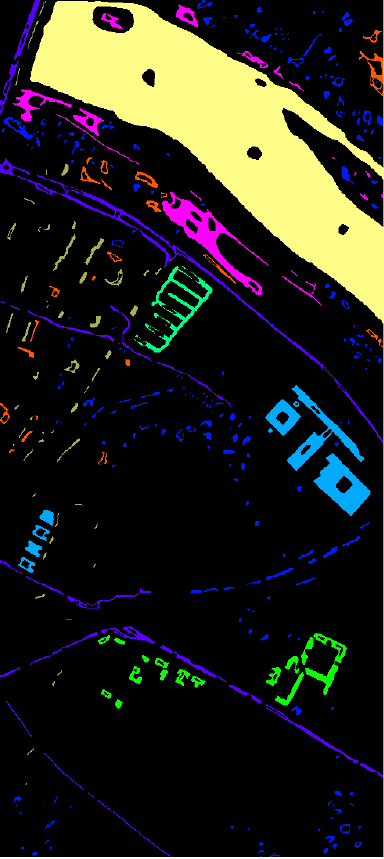}}
\centering \subfigure[]{
\includegraphics[width = 2.45in]{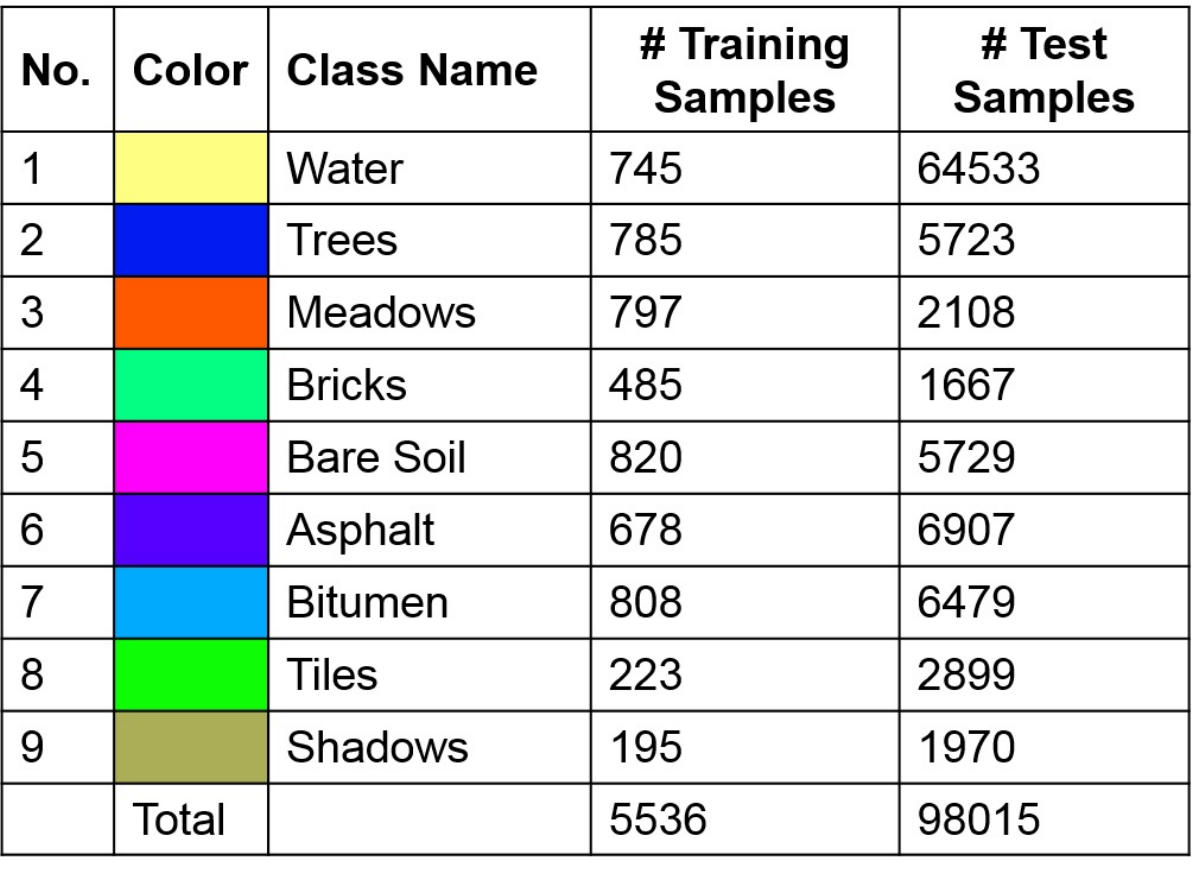}}
\caption{(a) False color composition of the Pavia Center scene. (b) Reference map containing 9 mutually exclusive land-cover classes.(c) Training samples. (d) Testing samples.(e)The numbers of the labeled samples.} \label{Pavia_data}
\end{figure}
\end{enumerate}

Three metrics are used for quantitative evaluation~\cite{cao2018hyperspectral}: overall accuracy (OA), average accuracy (AA) and the kappa coefficient~$\kappa$. OA is computed as the percentage of correctly classified test pixels, AA is the mean of the percentage of correctly classified pixels for each class, and $\kappa$ involves both omission and commission errors and gives a good representation of the the overall performance of the classifier.

For our method, the values of the kernel width $\delta$ in PRI were tuned around the multivariate Silverman's rule-of-thumb~\cite{silverman1986density}: $(\frac{4}{d+2})^{\frac{1}{d+4}} s^{\frac{-1}{4+d}}\sigma_1\leq\delta\leq (\frac{4}{d+2})^{\frac{1}{d+4}} s^{\frac{-1}{4+d}} \sigma_2$, where $s$ is the sample size, $d$ is the variable dimensionality, $\sigma_1$ and $\sigma_2$ are respectively the smallest and the largest standard deviation among each dimension of the variable. For example, in Indian Pines data set, the estimated range in the $5$-th layer corresponds to $[0.05, 0.51]$, and we set kernel width to $0.4$. On the other hand, the PRI in each layer is optimized with $\tau=3$ iterations, which has been observed to be sufficient to provide desirable performance.

\subsection{Parameter Analysis}

\subsubsection{Effects of parameter $\beta$ in PRI} \label{sec:beta_effect}
\begin{figure}[htbp]
\centering\subfigure[]{
\includegraphics[width = 1.8in]{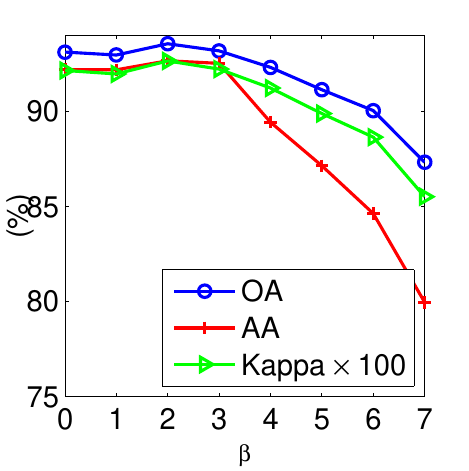}}
\centering \subfigure[]{
\includegraphics[width = 1.8in]{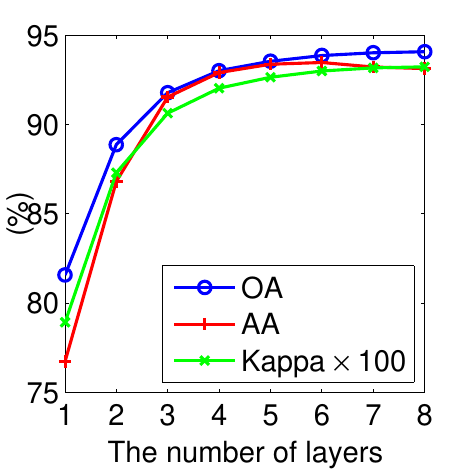}}
\caption{(a) Quantitative evaluation with different values of $\beta$. (b) Quantitative evaluation with different number of layers.} \label{beta1}
\end{figure}

The parameter $\beta$ in PRI balances the trade-off between the regularity of extracted representation and its discriminative power to the given data.
We illustrate the values of OA, AA, and $\kappa$ for MPRI with respect to different values of $\beta$ in Fig.~\ref{beta1}(a). As can be seen, these quantitative values are initially stable, but decrease when $\beta\geq3$. Moreover, the value of AA drops more drastically than that of OA or $\kappa$. A likely interpretation is that when training samples are limited, many classes have only a few labeled samples ($\sim1$ for minority classes, such as Oats, Grass-pasture-mowed, and Alfalfa). An unreasonable value of $\beta$ may severely influence the classification accuracy in these classes, hereby decreasing AA at first.

\begin{figure}[!htbp]
\centering \subfigure[]{
\includegraphics[height = 1.1in]{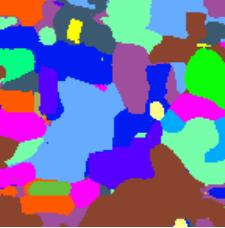}}
\centering \subfigure[]{
\includegraphics[height = 1.1in]{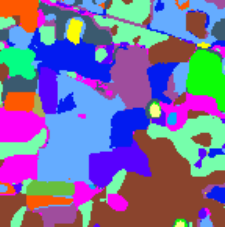}}
\centering \subfigure[]{
\includegraphics[height = 1.1in]{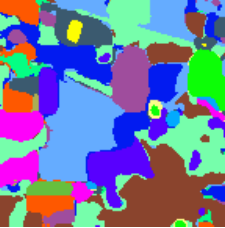}}
\centering \subfigure[]{
\includegraphics[height = 1.1in]{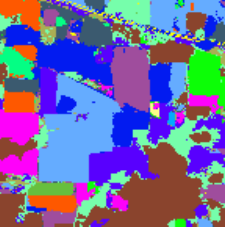}}\\
\centering \subfigure[]{
\includegraphics[height =1.1in]{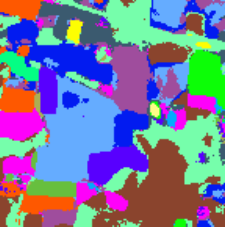}}
\centering \subfigure[]{
\includegraphics[height = 1.1in]{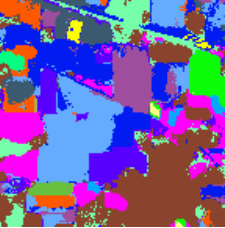}}
\centering \subfigure[]{
\includegraphics[height = 1.1in]{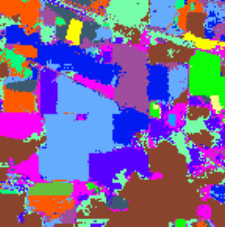}}
\centering \subfigure[]{
\includegraphics[height = 1.1in]{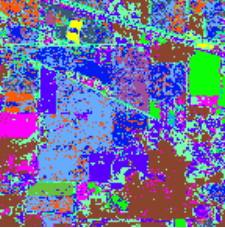}}
\caption{Classification maps of MPRI with (a) $\beta=0$; (b) $\beta=1$; (c) $\beta=2$; (d) $\beta=3$; {(e) $\beta=4$; (f) $\beta=5$;} (g) $\beta=6$; and (h) $\beta=100$.} \label{beta2}
\end{figure}

The corresponding classification maps are shown in Fig.~\ref{beta2}. It is obviously that, the smaller the $\beta$, the more smooth results achieved by MPRI. This is because large $\beta$ encourages a small divergence between the extracted representation and the original HSI data. For example, in the scenario of $\beta=0$, PRI clusters both spectral and spatial structures into a single point (the data mean) that has no discriminative power. By contrast, in the scenario of $\beta\rightarrow\infty$, the extracted representation gets back to the HSI data itself (to minimize their divergence) such that PRI will fit all noisy and irregular structures.

From the above analysis, extremely large and small values of $\beta$ are not interesting for classification of HSI. Moreover, the results also suggest that $\beta\in[2, 4]$ is able to balance a good trade-off between preserving relevant spatial information (such as edges in classification maps) and filtering out unnecessary one. Unless otherwise specified, the PRI mentioned in the following experiments uses three different values of $\beta$, i.e., $\beta=2$, $\beta=3$, and $\beta=4$. The final representation of PRI is formed by concatenating representations obtained from each $\beta$.

\subsubsection{Effects of the number of layers}
\begin{figure}
\centering \subfigure[]{
\includegraphics[height = 1.1in]{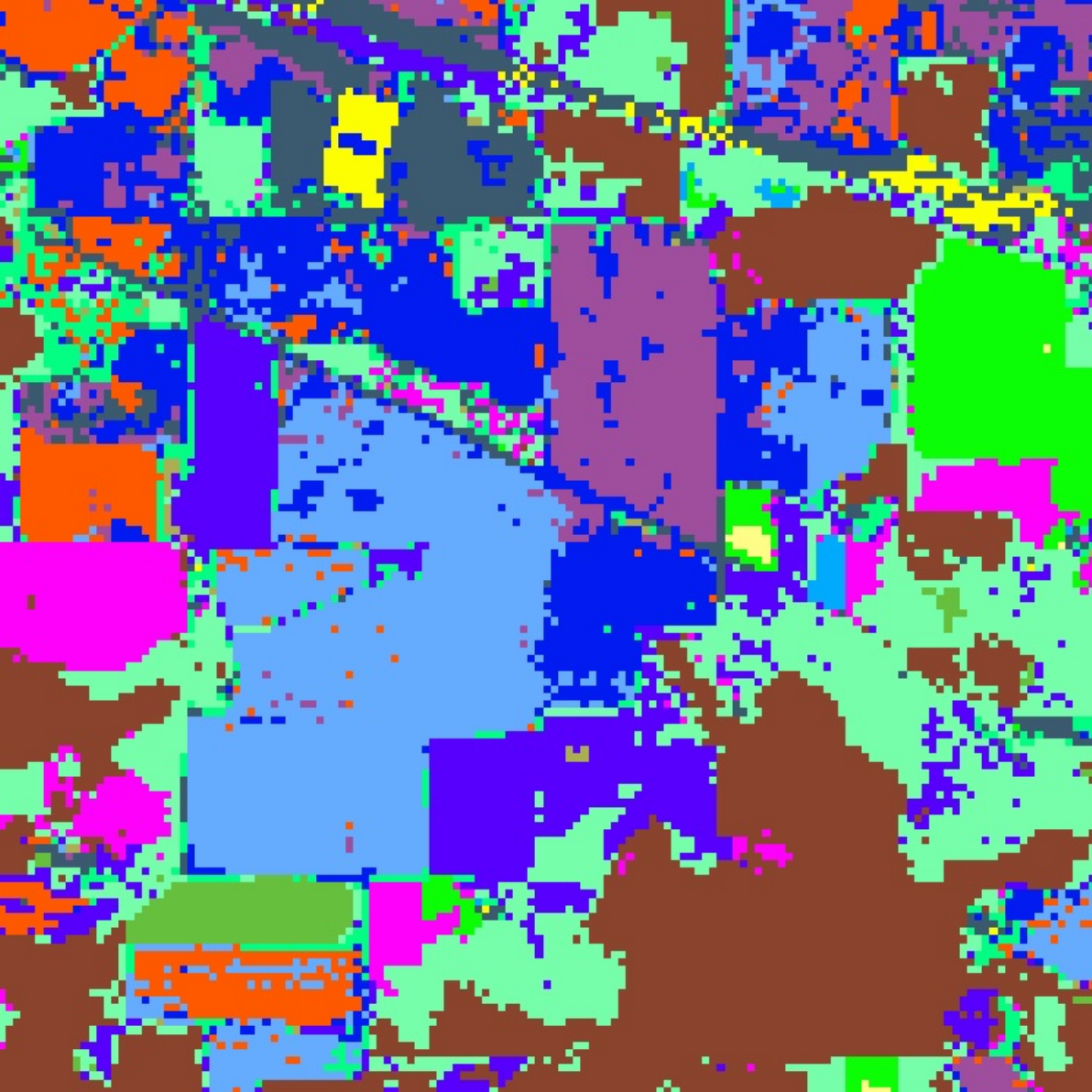}}
\centering \subfigure[]{
\includegraphics[height = 1.1in]{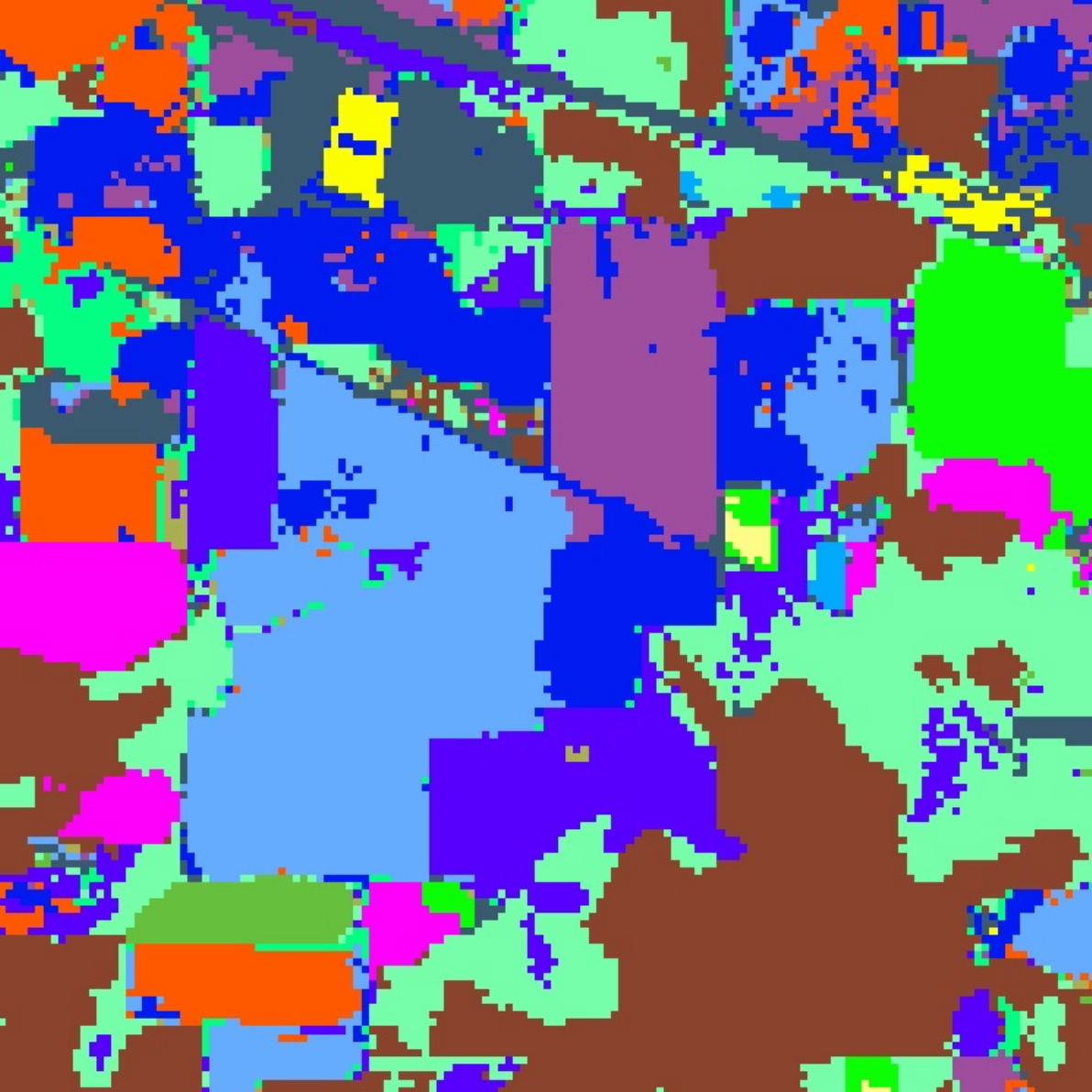}}
\centering \subfigure[]{
\includegraphics[height = 1.1in]{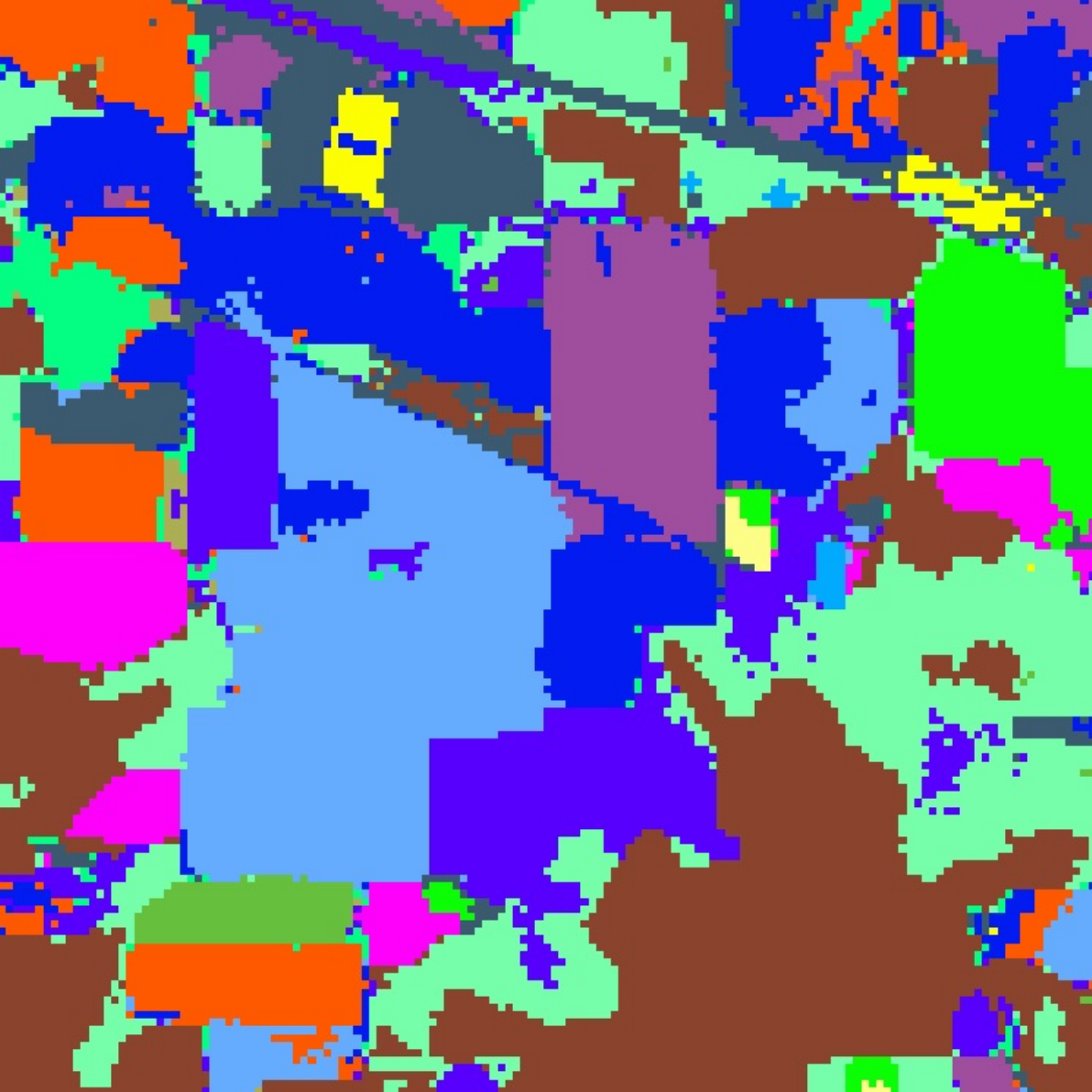}}
\centering \subfigure[]{
\includegraphics[height =  1.1in]{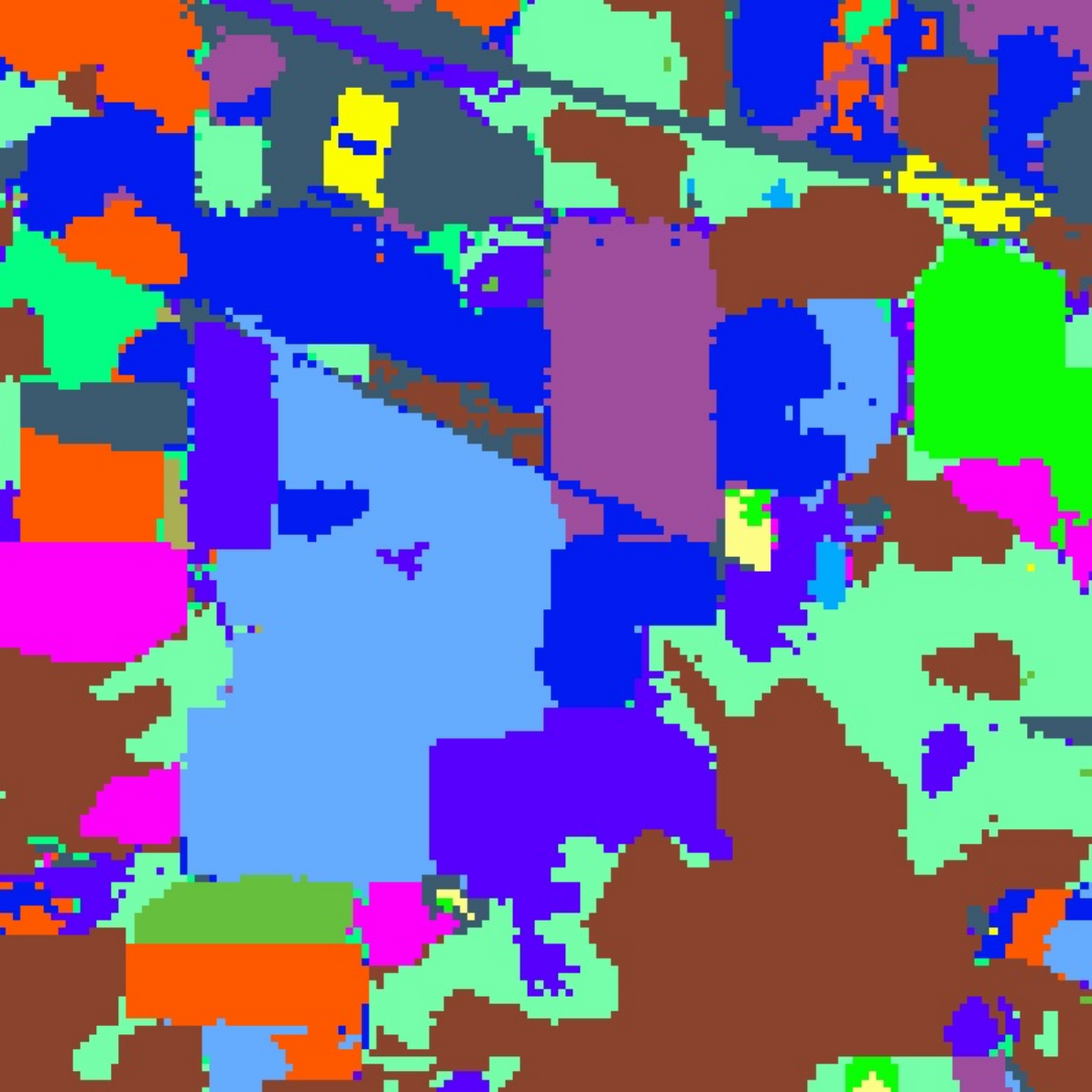}}\\
\centering \subfigure[]{
\includegraphics[height =  1.1in]{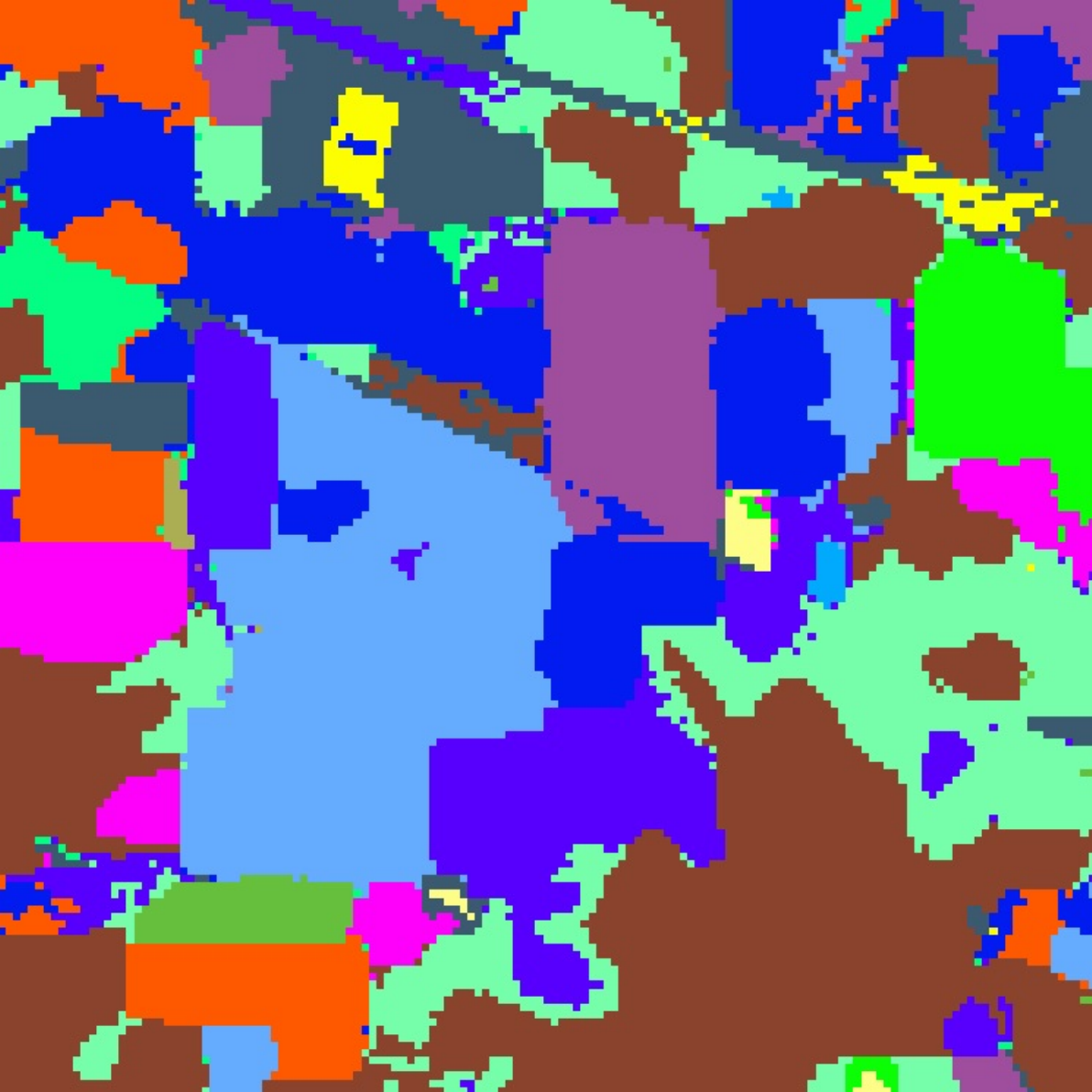}}
\centering \subfigure[]{
\includegraphics[height =  1.1in]{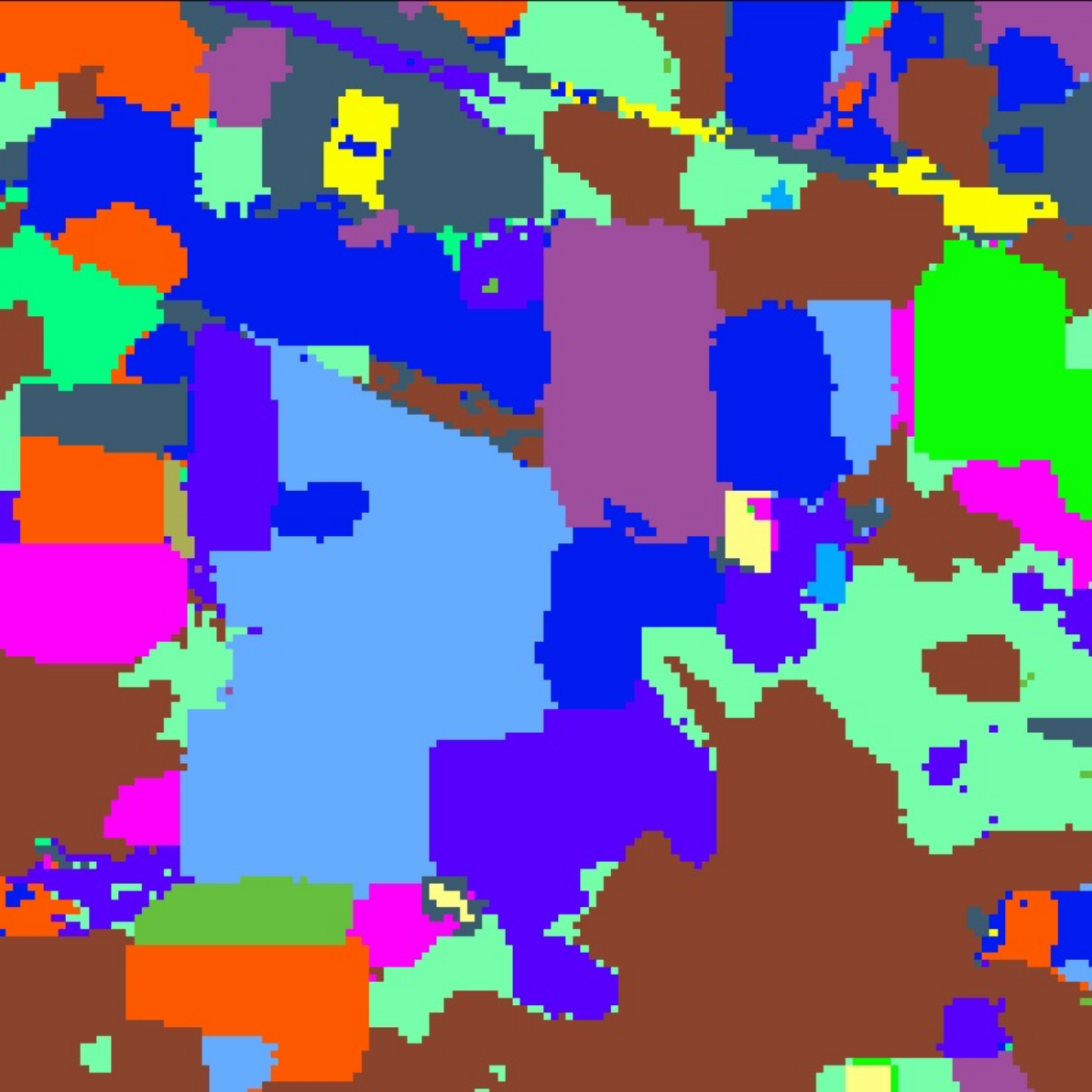}}
\centering \subfigure[]{
\includegraphics[height = 1.1in]{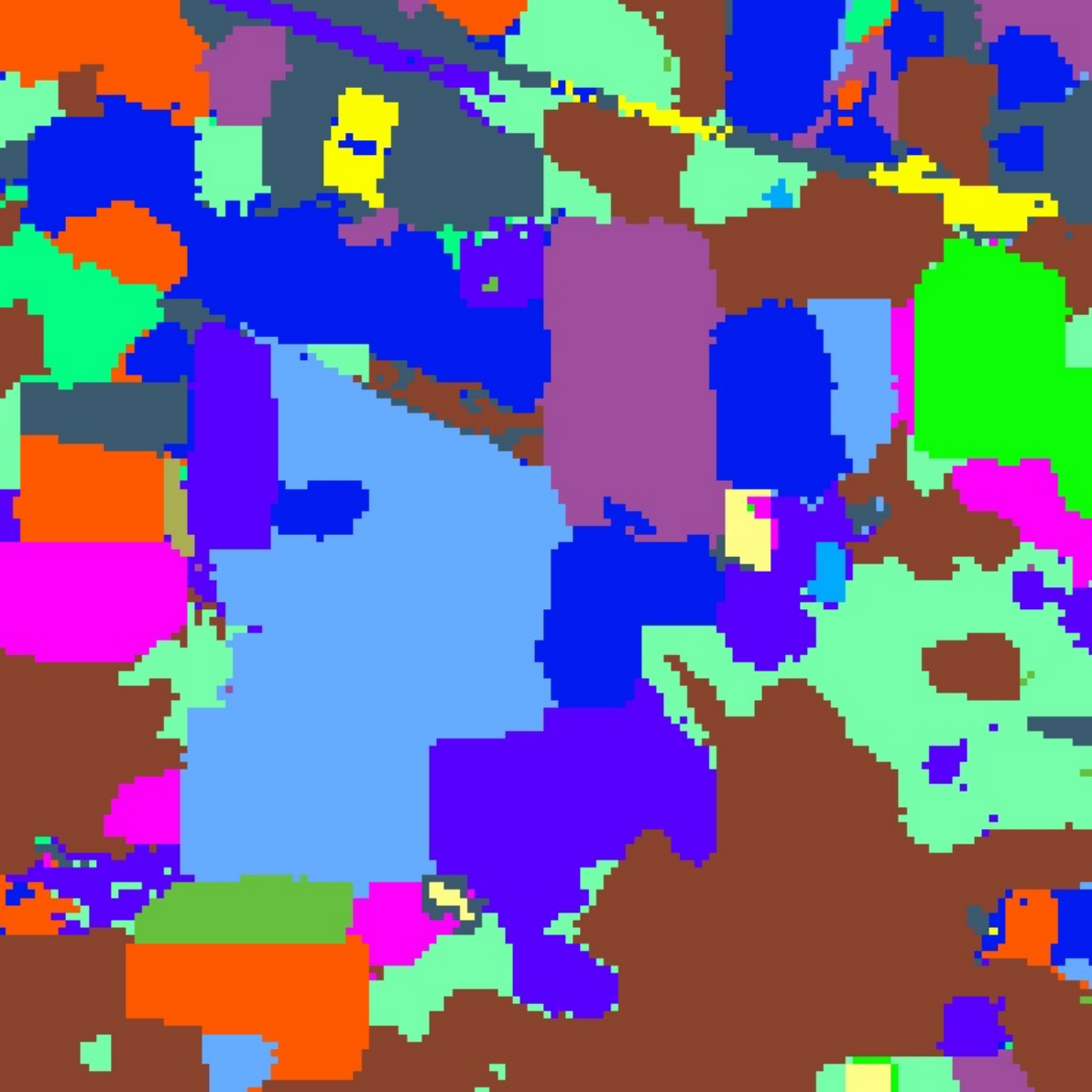}}
\centering \subfigure[]{
\includegraphics[height = 1.1in]{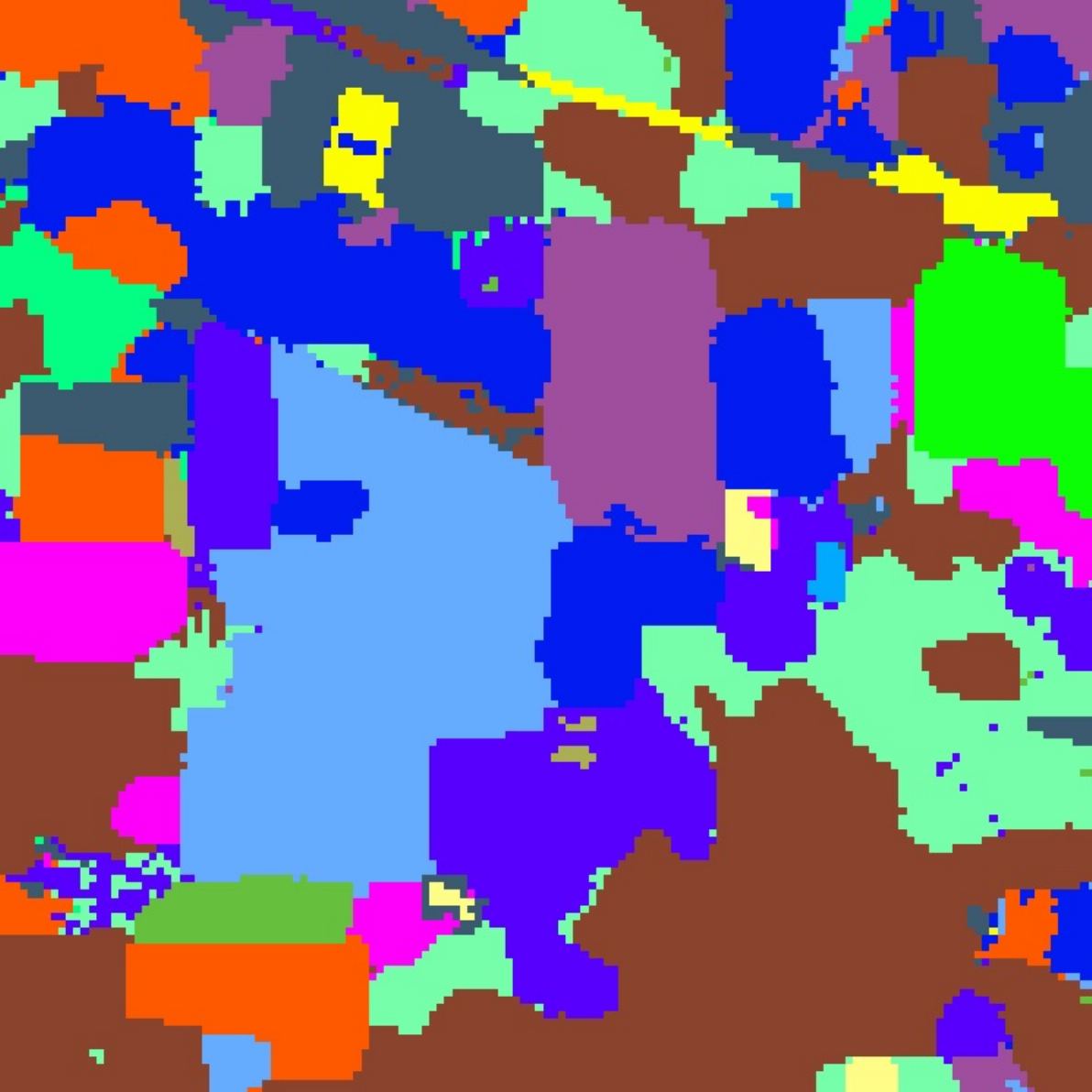}}
\caption{Classification maps of MRPI with (a) $1$ layer; (b) $2$ layers; (c) $3$ layers; (d) $4$ layers; (e) $5$ layers; (f) $6$ layers; (g) $7$ layers; and (h) $8$ layers.} \label{layer2}
\end{figure}

We then illustrate the values of OA, AA and $\kappa$ for MPRI with respect to different number of layers in Fig.~\ref{beta1}(b). The corresponding classification maps are shown in Fig.~\ref{layer2}.
Similar to existing deep architectures, stacking more layers (in a reasonable range) can increase performance. If we keep the input data size the same, more layers (beyond a certain layer number) will not increase the performance anymore and the classification maps become over-smooth. This work uses a $5$-layer MPRI because it provides favorable visual and quantitative results.

\subsubsection{Effects of the classifier}
MPRI uses the basic KNN for classification and sets $k=1$ throughout this work. The purpose is to validate the discriminative power of the spectral-spatial features extracted by multiple layers of PRI. To further confirm the superiority of our MPRI is independent to the used classifier, we also evaluate the performances of MPRI and other three feature extraction methods for HSI classification (EPF \cite{X.Kang14}, SADL \cite{A.Soltani-Farani15}, PCA-EPF \cite{kang2017pca}) and use a kernel SVM as the baseline classifier. The kernel size $\sigma$ is tuned by cross-validation from the set $\{0.0001, 0.001, 0.01, 0.1, 1, 10, 100\}$. The best performance is summarized in Table~\ref{Indian_accuracy_SVM}. As can be seen, KNN and SVM always lead to comparable results. Our MPRI is consistently better than other competitors, regardless of the used classifier.

\begin{table}[htbp]
\caption{OAs (\%) of different methods using KNN and SVM. 2\% of labeled samples per class were randomly selected for training. The best performances are marked in bold.}
\begin{center}{\begin{tabular}{ccccc}
\toprule
Classifier & EPF&SADL&PCA-EPF&MPRI\\
\midrule
KNN ($k=1$) &77.02&{87.48}&{{90.03}}& {\bf 94.00}\\
SVM ($\sigma=0.1$) &78.44&{88.19}&{91.77}& {\bf 93.11}\\
\bottomrule
\end{tabular}}
\end{center}
\label{Indian_accuracy_SVM}
\end{table}

\subsection{Evaluation on component-wise contributions}

\begin{table}[htbp]
\caption{Quantitative evaluation of our MPRI (the last row) and its degraded baseline variants in terms of OA, AA, and $\kappa$. ``$\checkmark$" denotes the model contains the corresponding module. The best two performances are marked in bold and underlined, respectively.}
\begin{center}
{\begin{tabular}{c c c | c c c c}
\toprule
\emph{Multi-layer} & \emph{Multi-scale} & \emph{Multi-$\beta$} & OA & AA & $\kappa$\\
\midrule
 & & &76.25&72.74&0.729\\
\checkmark & & &80.86&79.33&0.7814\\
 & \checkmark & &81.28&76.25&0.786\\
 & & \checkmark &76.61&73.09&0.733\\
 & \checkmark & \checkmark & 83.30&77.89&0.809\\
\checkmark & & \checkmark & 86.94&85.35&0.851\\
\checkmark & \checkmark & & \underline{92.80}&\underline{91.16}&\underline{0.918}\\
\checkmark & \checkmark & \checkmark & \bf{94.00} & \bf{91.20} & \bf{0.932}\\
\bottomrule
\end{tabular}}
\end{center}
\label{BV}
\end{table}
Before systematically evaluating the performance of MPRI, we first compare it with its degraded baseline variants to demonstrate the component-wise contributions to the performance gain. The results are summarized in Table~\ref{BV}. As can be seen, models that only consider one attribute (i.e., \emph{multi-layer}, \emph{multi-scale} and \emph{multi-$\beta$}) improve the performance marginally. Moreover, it is interesting to find that \emph{multi-layer} and \emph{multi-scale} play more significant roles than \emph{multi-$\beta$}. One possible reason is that the representations learned from different $\beta$ contain redundant information with respect to class labels. However, either the combination of \emph{multi-layer} and \emph{multi-$\beta$} or the combination of \emph{multi-scale} and \emph{multi-$\beta$} can obtain remarkable improvements. Our MPRI performs the best as expected. This result indicates that \emph{multi-layer}, \emph{multi-scale} and \emph{multi-$\beta$} are essentially important for the problem of HSI classification.

\subsection{Comparison with state-of-the-art methods}

Having illustrated component-wise contributions of MPRI, we compare it with several state-of-the-art methods, including EPF \cite{X.Kang14}, MPM-LBP \cite{J.Li13}, SADL \cite{A.Soltani-Farani15}, MFL \cite{li2015multiple}, PCA-EPF \cite{kang2017pca}, HIFI \cite{pan2017hierarchical}, hybrid spectral convolutional neural network (HybridSN) \cite{roy2019hybridsn}, similarity-preserving deep features (SPDF) \cite{fang2019deep},  convolutional neural network with Markov random fields (CNN-MRF) \cite{cao2018hyperspectral},  local covariance matrix representation (LCMR) \cite{fang2018new}, and random patches network (RPNet) \cite{xu2018hyperspectral}.



\begin{table}[htbp]
\caption{Classification accuracies (\%) of different methods on Indian Pines data set. 2\% of labeled samples per class were randomly selected for training. The best two performances are marked in bold and underlined, respectively.}
\begin{center}{\begin{tabular}{lp{0.5cm}p{0.6cm}p{0.6cm}p{0.6cm}p{0.6cm}p{0.6cm}p{0.6cm}p{0.6cm}p{0.6cm}p{0.6cm}p{0.6cm}p{0.6cm}}
\toprule
Class & EPF&MPM-LBP&SADL&MFL&PCA-EPF&HIFI&Hybr-idSN&SPDF&CNN-MRF&RPNet&MPRI\\
\midrule
1&{10.00}&51.73& {75.77} &{66.92} &{\bf 98.76}&{ {\underline {87.33}}}&{17.78}&40.44& {46.67}&{14.67}&{74.44}\\
2&74.41&80.98&{83.89} &{80.58} &{\underline {88.87}}&{ 88.34}&70.29&81.66&{70.33} &{86.90}&{\bf 92.48}\\
3&89.35&68.46&{81.57}&{76.41} &88.15&{\underline{89.00}}&71.87&73.01&{67.75}&{69.43}&{\bf 95.68}\\
4&79.01&44.76&{64.15}&{ 26.46} &{\bf 87.08}&\underline{81.85}&70.69&53.88&{43.36} &{26.21}&{72.67}\\
5&{\bf 95.15}&80.86&{90.12}&{80.76}&\underline{94.78}&{83.97}&83.30&77.02&{78.94}&{83.62}&{89.98}\\
6&73.93&{\bf 97.87}&{97.34}&{91.20} &92.81&{96.90}&\underline {97.76}&85.94&{85.69}&{90.69}&{96.91}\\
7&{50.00}&29.20&{{\bf 100.0}}&{67.20} &80.30&{96.30}&48.15&40.74&{{36.30}}&{70.74}&\underline {96.67}\\
8&86.57&98.37&{{99.65}}&{96.41} &\underline{99.83}&{98.57}&99.57&88.63&{{92.46}} &{84.81}&{\bf 100.0}\\
9&	10.00&40.53&\underline{93.16}&{ 32.11} &76.45&{ \bf 96.84}&	0.000&40.53&{69.47} &{41.58}&{ 81.05}\\
10&77.92&75.79&{85.79}&{82.59} &\underline{90.10}&{88.80}&79.22&70.41&{63.03}&{78.68}&{\bf 92.68}\\
11&71.76&90.25&{91.55}&{ 90.50} &{92.65}&{90.19}&91.77&90.89&{87.88} &{\bf 95.92}&\underline{94.90}\\
12&74.80&79.53& {66.27}&{70.70} &{86.78}&\underline{87.30}&54.73&69.98& {59.52} &{48.43}&{ \bf87.93}\\
13&{ 98.02}&{\bf 99.47}&{98.55}&{96.28} &96.40&\underline{99.30}&78.11&80.50&{71.05} &{89.10}&{98.40}\\
14&91.21&96.36&{96.34} &{96.47} &97.72&\underline{98.24}&96.53&92.15&{95.05}  &{92.17}&\bf{98.59}\\
15&78.84&63.17&{84.78}&{74.11} &{ \bf 94.11}&{81.83}&63.76&72.09&{57.27} &{55.63}&\underline{93.89}\\
16& 87.15&50.54&{{ 90.32}}&{73.01} &89.67&{\bf 98.24}&75.82&46.15&{80.77} &{29.67}&\underline{92.97}\\
\midrule
OA &78.44&83.84&{88.19}&{84.32}&{\underline{91.77}}&{ 90.87}&	82.22&81.41&{77.39}&{81.80}& {\bf 94.00}\\
AA &71.76&71.74&{87.45}&{75.11}&90.90&{\bf 91.44}&	68.71&69.00&{69.10}&{66.14}& {\underline{91.20}}\\
$\kappa$&0.750&0.814&{0.865}&{0.821} &{\underline{0.906}}&{0.896}&0.797&0.786&{0.740}&{0.788}&{\bf 0.932}\\
\bottomrule
\end{tabular}}
\end{center}
\label{Indian_accuracy_o}
\end{table}

\begin{table}[htbp]
\caption{Classification accuracies (\%) of different methods on Pavia University data set. 1\% of labeled samples per class were randomly selected for training. The best two performances are marked in bold and underlined, respectively.}
\begin{center}{\begin{tabular}{lp{0.5cm}p{0.6cm}p{0.6cm}p{0.6cm}p{0.6cm}p{0.6cm}p{0.6cm}p{0.6cm}p{0.6cm}p{0.6cm}p{0.6cm}p{0.6cm}}
\toprule
Class & EPF&MPM-LBP&SADL&MFL&PCA-EPF&HIFI&Hybr-idSN&SPDF&CNN-MRF&RPNet&MPRI\\
\midrule
1&93.05&97.38&{92.66}&{\bf 98.05}&96.56&{92.37}&90.78&96.10&{93.79}&{91.40}&\underline{97.75}\\
2&95.37&99.45&{98.92}&{99.61}&98.67&{98.54}&\underline{99.84}&99.25&{98.51}&{97.94}&{\bf 99.88}\\
3&{\bf 96.87}&79.02&{74.65}&{74.35}&87.43&{80.40}&71.80&\underline{92.98}&{65.04}&{86.48}&{89.52}\\
4&{\bf 99.46}&{ 90.62}&{93.50}&{89.78}&\underline{98.34}&{81.61}&73.95&{83.54}&{95.95}&{91.23}&{94.05}\\
5&98.09&{97.52}&{\underline{99.38}}&{98.45}&{\bf 99.44}&{96.39}&98.05&{96.06}&{{96.66}}&{95.51}&{98.87}\\
6&{ 98.52}&	93.37&{94.57}&{95.04 }&\underline{99.00}&{ 90.49}&{\bf 99.86}&95.07&{93.15}&{93.83}&{96.26}\\
7&\underline{99.54}&82.95&{77.26}&{94.12}&94.29&{87.67}&{\bf 99.62}&83.30&{64.00}&{81.66}&{99.02}\\
8&87.63&	91.52&{77.73}&\underline{93.11}&\underline{92.22}&{89.71}&92.24&{\bf 95.14}&{80.21}&{87.40}&{91.69}\\
9&96.88&98.94&{\underline{99.14}}&{91.54}&98.36&{\bf99.92}&59.23&55.39&{{93.55}}&{96.03}&{ 97.18}\\
\midrule
OA&95.06&	95.42&{93.28}&{ 95.83}&\underline{97.11}&{93.40}&	93.59&94.92&{92.51}&{93.87}&{\bf 97.31}\\
AA&	\underline{96.16}&{92.31}&{89.76}&{ 92.67}&{\bf 96.21}&{90.78}&	87.26&{88.54}&{86.76}&{91.28}&{96.03}\\
$\kappa$&		0.935&	0.940&	{0.912}&{0.945}&\underline{0.962}&{0.912} &	0.932&	0.940&	{90.07}&{0.918} & {\bf 0.965}\\
\bottomrule
\end{tabular}}
\end{center}
\label{PaviaU_accuracy}
\end{table}

\begin{table}[htbp]
\caption{Classification accuracies (\%) of different methods on Pavia Center data set with fixed training and testing split. The best two performances are marked in bold and underlined, respectively.}
\begin{center}{\begin{tabular}{lp{0.5cm}p{0.6cm}p{0.6cm}p{0.6cm}p{0.6cm}p{0.6cm}p{0.6cm}p{0.6cm}p{0.6cm}p{0.6cm}p{0.6cm}p{0.6cm}}
\toprule
Class & EPF&MPM-LBP&SADL&MFL&PCA-EPF&HIFI&Hybr-idSN&SPDF&CNN-MRF&RPNet&MPRI\\
\midrule
{1}&{{\bf 100.0}}&{ 99.10}&{{98.79}}& {99.43 }&{\bf 100.0}& \underline{99.57} &97.57&97.45&95.84&94.40&{\bf 100.0}\\
{2}&		{{\bf 99.39}}&{94.57}&{92.66}&{87.02 }&97.21&{92.63}&90.21&89.15&\underline{98.03}& 90.58&{97.76}\\
{3}&	{86.78}& {95.83}&{96.63}&{95.83 }&88.49&{96.16}&\underline{99.34}&98.29&86.01& 99.15&{\bf 99.86}\\
{4}&		{98.28}&{81.94}&{ 98.32}& \underline{99.46 }&97.74&{93.70}&97.78&59.87&99.22& 80.50&{\bf 100.0}\\
{5}&{96.64}&{97.75}&{97.91}&\underline{99.43}&{\bf 99.54}&{98.32}&98.22&92.15&98.22&99.09 &{98.55}\\
{6}&{77.48}&{95.21}& {95.51}&{95.05 }&93.93&\underline{98.39}&99.48&97.47&92.24& 96.24&{{\bf 99.83}}\\
{7}&{95.29}&{94.51}&{98.24}&{96.39 }&{\bf 99.84}&{94.00}&{\bf 99.51}&87.16&98.69& 90.83&{{ 97.85}}\\
{8}&{{\bf 100.0}}&{99.62}& {98.76}&{99.34 }&98.61&{99.72}&\underline{99.97}&84.20&99.76&88.58 &{\bf 100.0}\\
{9}&{{\bf 100.0}}&{{\bf 100.0}}&{{\bf 100.0}}&{97.23}&{\bf 100.0}&\underline{99.95}&97.97&92.23&99.29&94.06 &{97.31}\\
\midrule
{OA}&{	97.08}&{97.85}& {98.08}&{98.01 }&\underline{99.01}&{98.48}&97.81&94.86&96.07& 94.03&{{\bf 99.57}}\\
{AA}&	{94.87} &{95.39}& {\underline{97.42}}&{96.47 }&97.26&{96.94}&97.82&88.66&96.36& 92.60&{\bf99.02}\\
{$\kappa$}&	{0.948}&{0.961}& {0.965}&{0.964 }&\underline{0.982}&{0.972}&0.961&0.908&0.931& 0.895&{{\bf 0.992}}\\
\bottomrule
\end{tabular}}
\end{center}
\label{Pavia_Center_accuracy_fixed}
\end{table}

Tables~\ref{Indian_accuracy_o}-\ref{Pavia_Center_accuracy_fixed} summarized quantitative evaluation results of different methods. For each method, we report its classification accuracy in each land cover category as well as the overall OA, AA and $\kappa$ values across all categories. To avoid biased evaluation, we average the results from $10$ independent runs (except for the Pavia Center data set, in which the training and testing samples are fixed). Obviously, MPRI achieves the best or the second best performance in most of items. These results suggest that MPRI is able to learn more discriminative spectral-spatial features than its counterparts using classical machine learning models.

\begin{figure}[htbp]
\centering \subfigure[]{
\includegraphics[width = 1.5in]{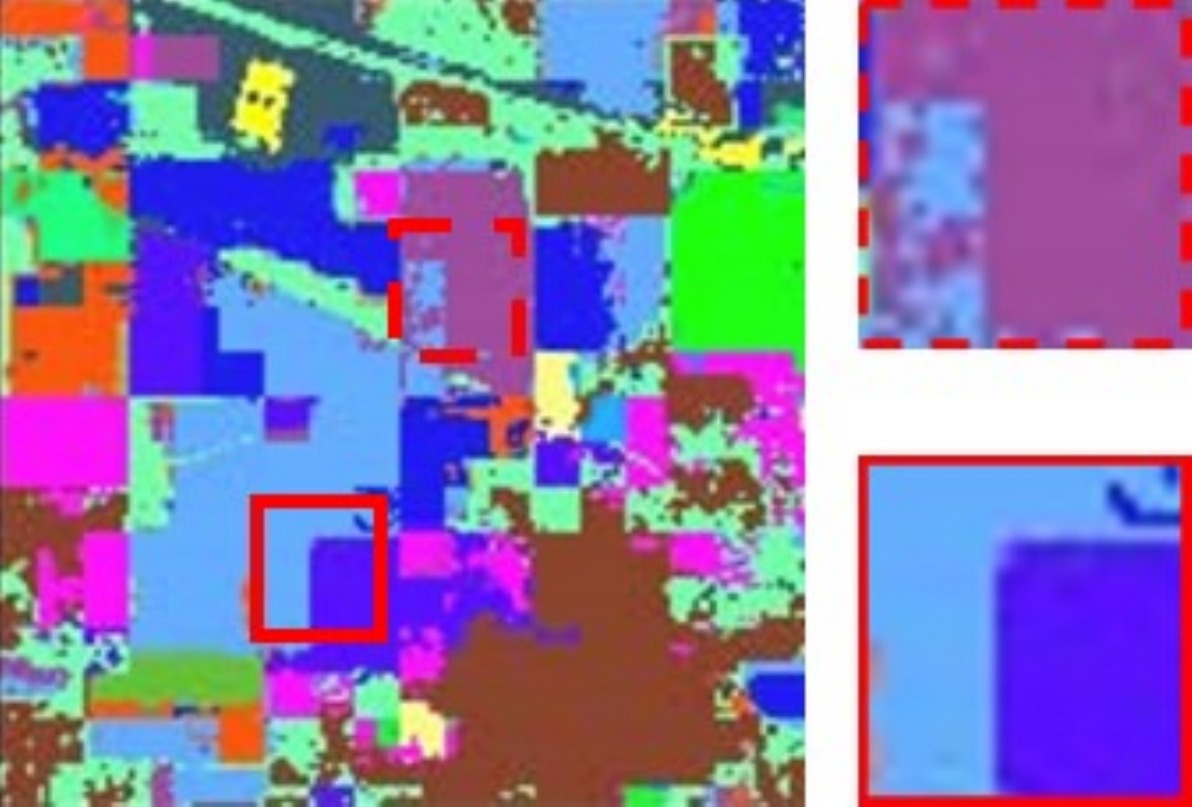}}
\centering \subfigure[]{
\includegraphics[width = 1.5in]{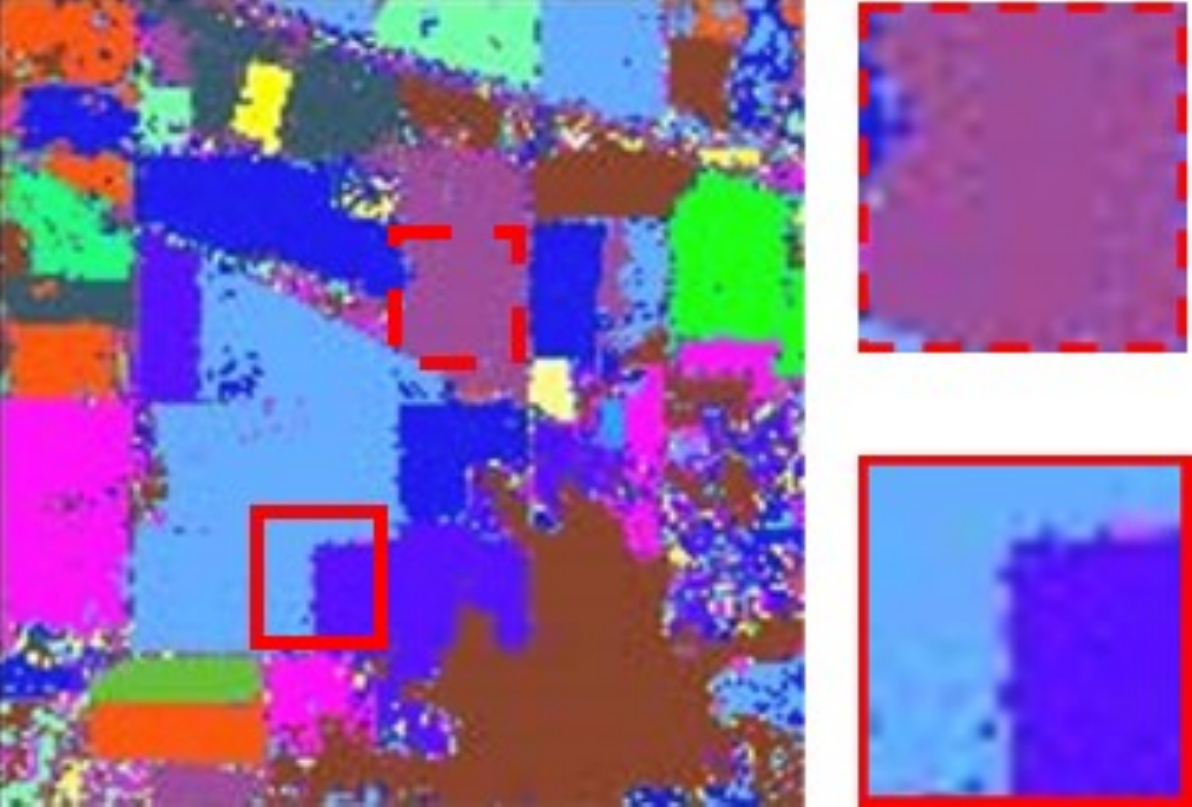}}
\centering \subfigure[]{
\includegraphics[width = 1.5in]{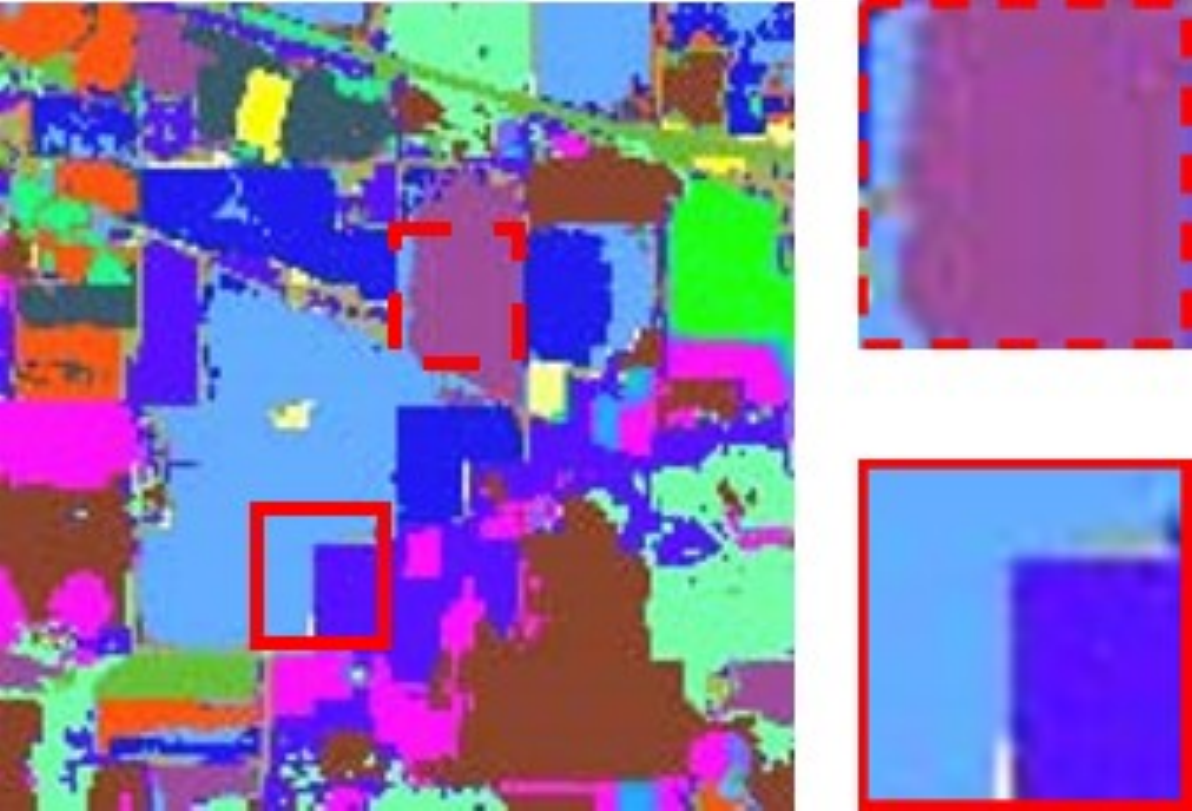}}\\
\centering \subfigure[]{
\includegraphics[width = 1.5in]{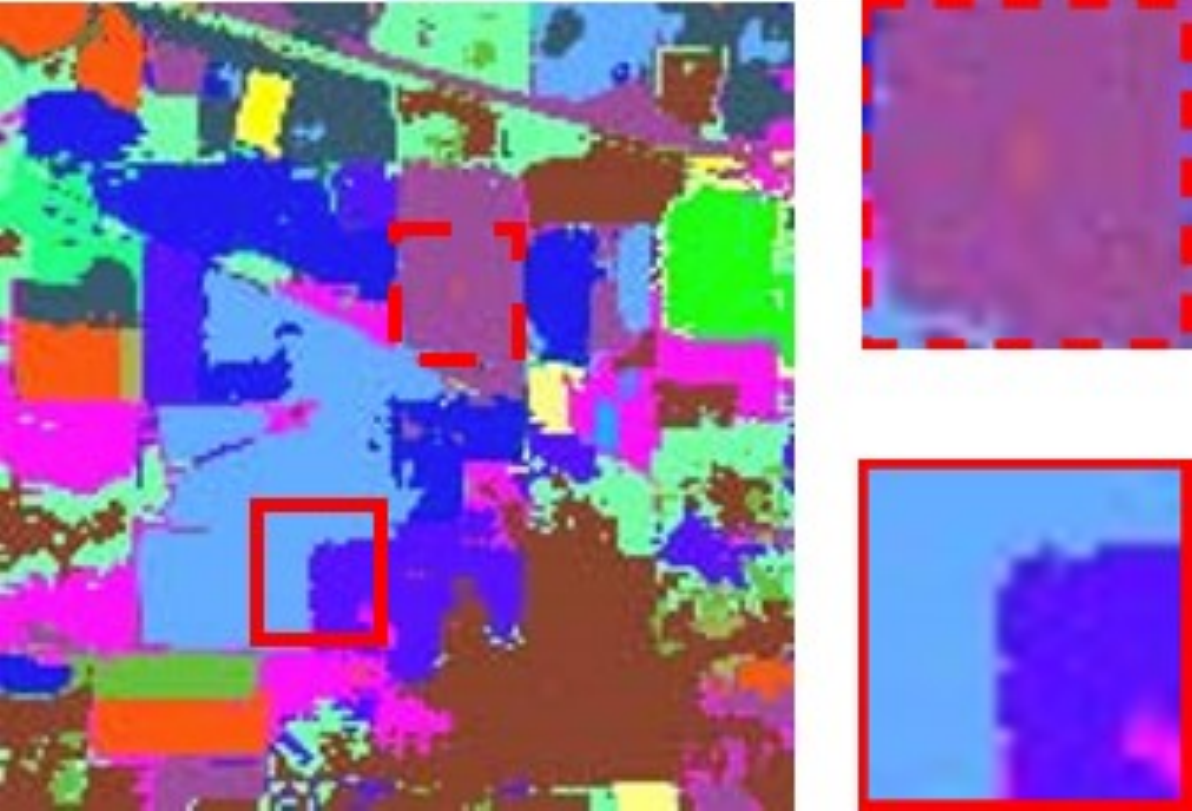}}
\centering \subfigure[]{
\includegraphics[width = 1.5in]{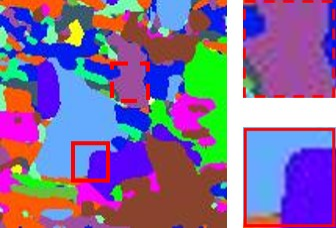}}
\centering \subfigure[]{
\includegraphics[width = 1.5in]{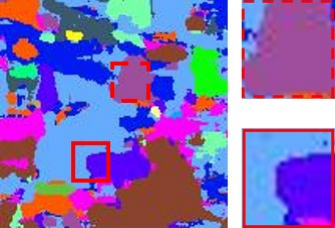}}\\
\centering \subfigure[]{
\includegraphics[width = 1.5in]{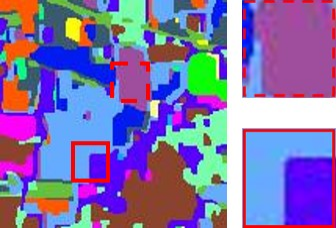}}
\centering \subfigure[]{
\includegraphics[width = 1.5in]{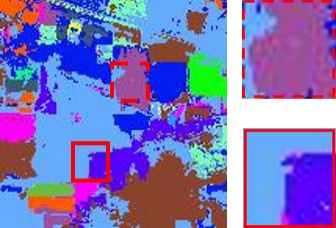}}
\centering \subfigure[]{
\includegraphics[width = 1.5in]{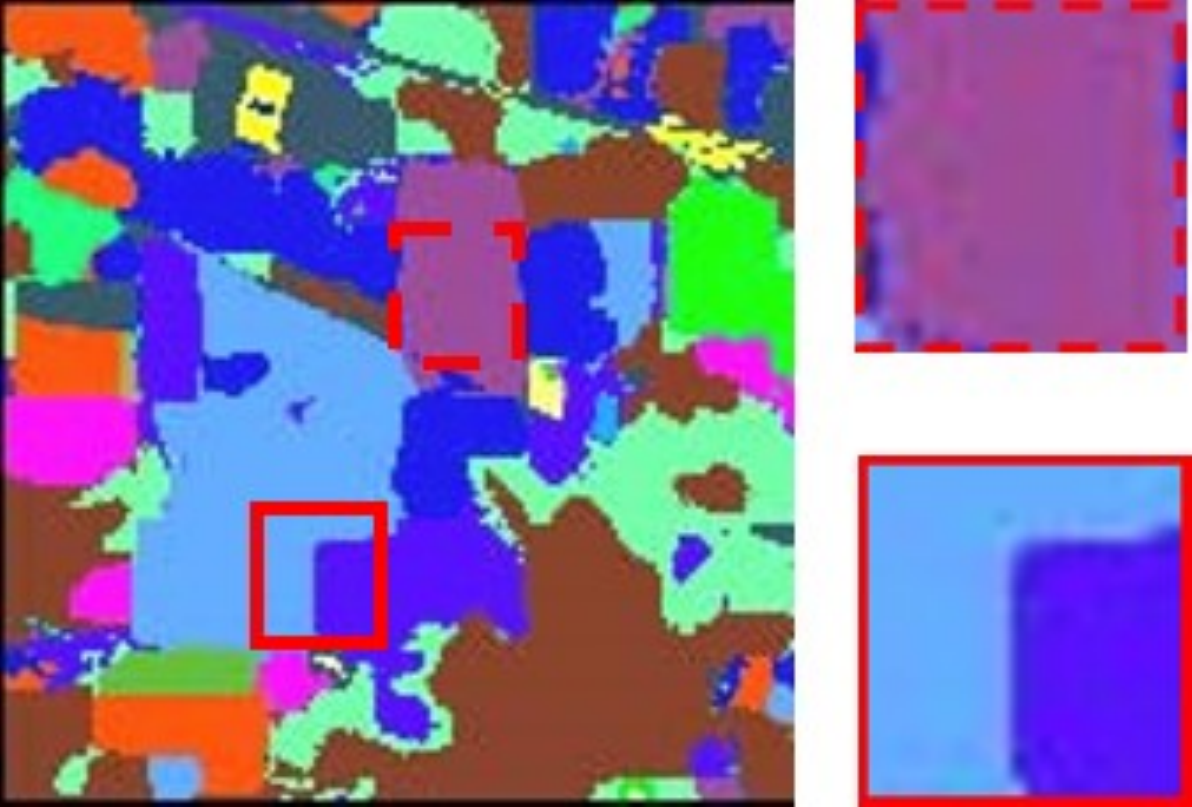}}
\caption{Classification maps on Indian Pines data set. (a) SADL; (b) MFL;  (c) PCA-EPF;  (d) HIFI;  (e) HybridSN; (f) SPDF; (g) CNN-MRF; (h) RPNet;  (i) MPRI.} \label{Indian Pines_Train}
\end{figure}

\begin{figure}[htbp]
\centering \subfigure[]{
\includegraphics[width = 1.5in]{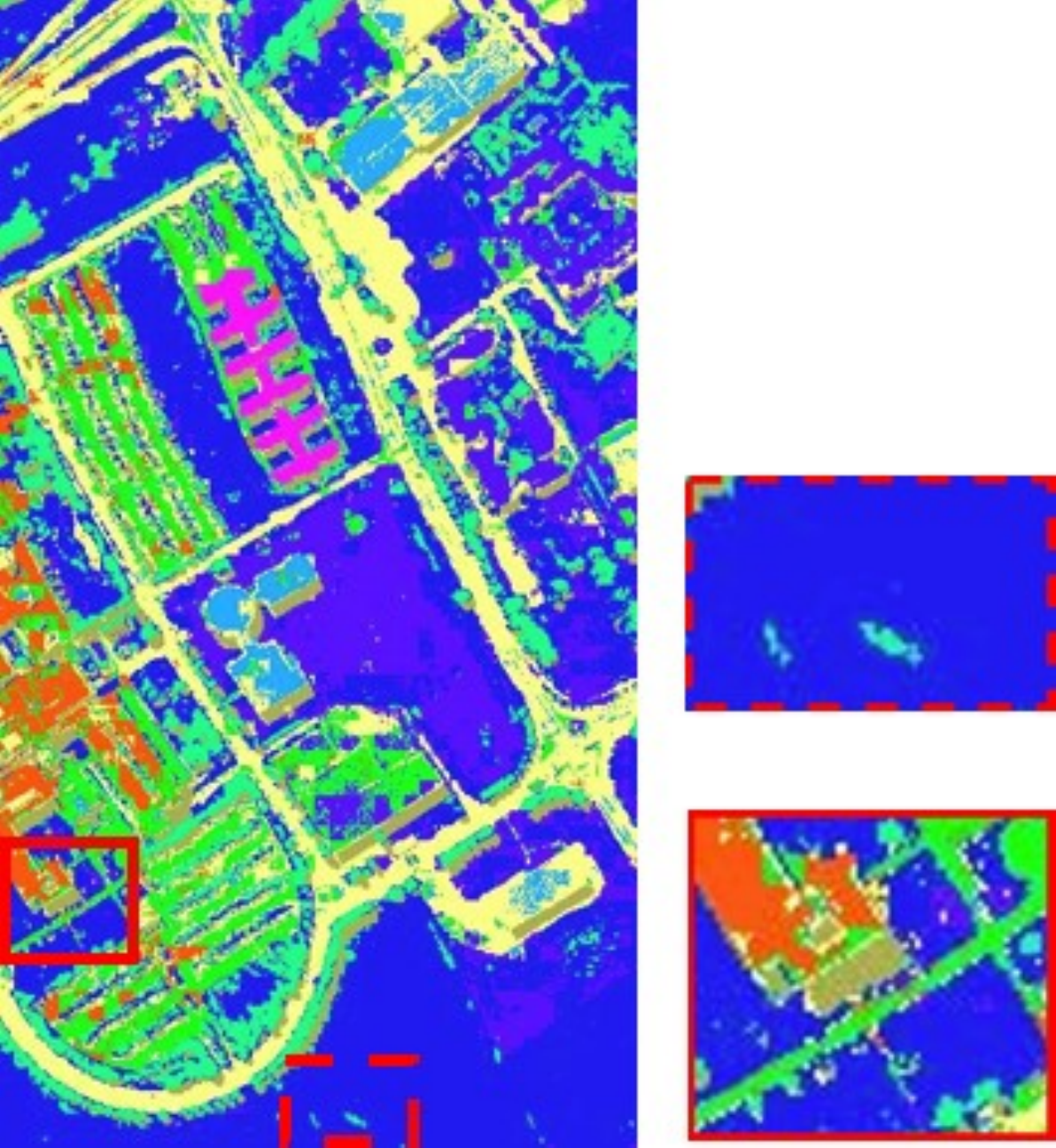}}
\centering \subfigure[]{
\includegraphics[width = 1.5in]{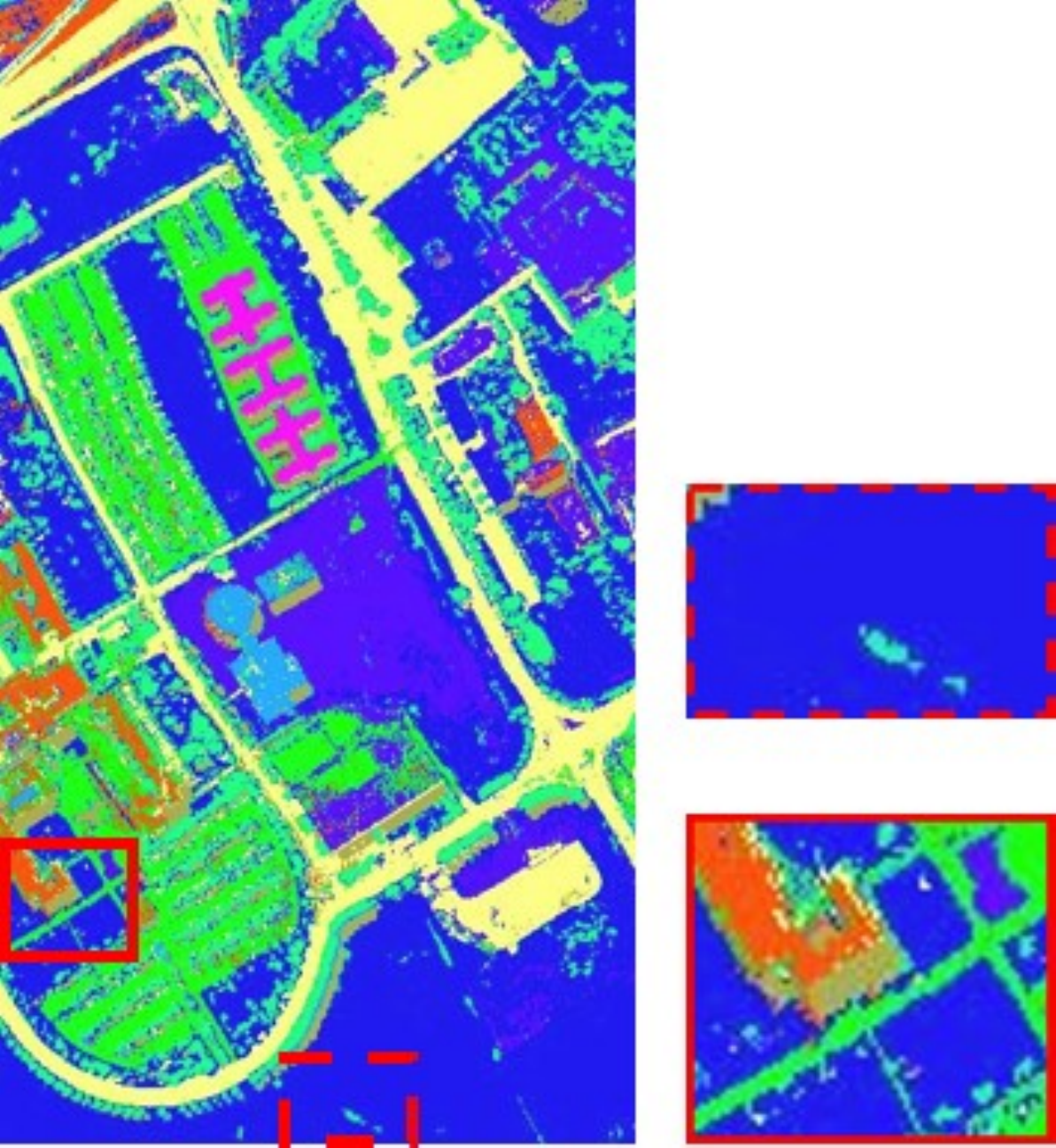}}
\centering \subfigure[]{
\includegraphics[width = 1.5in]{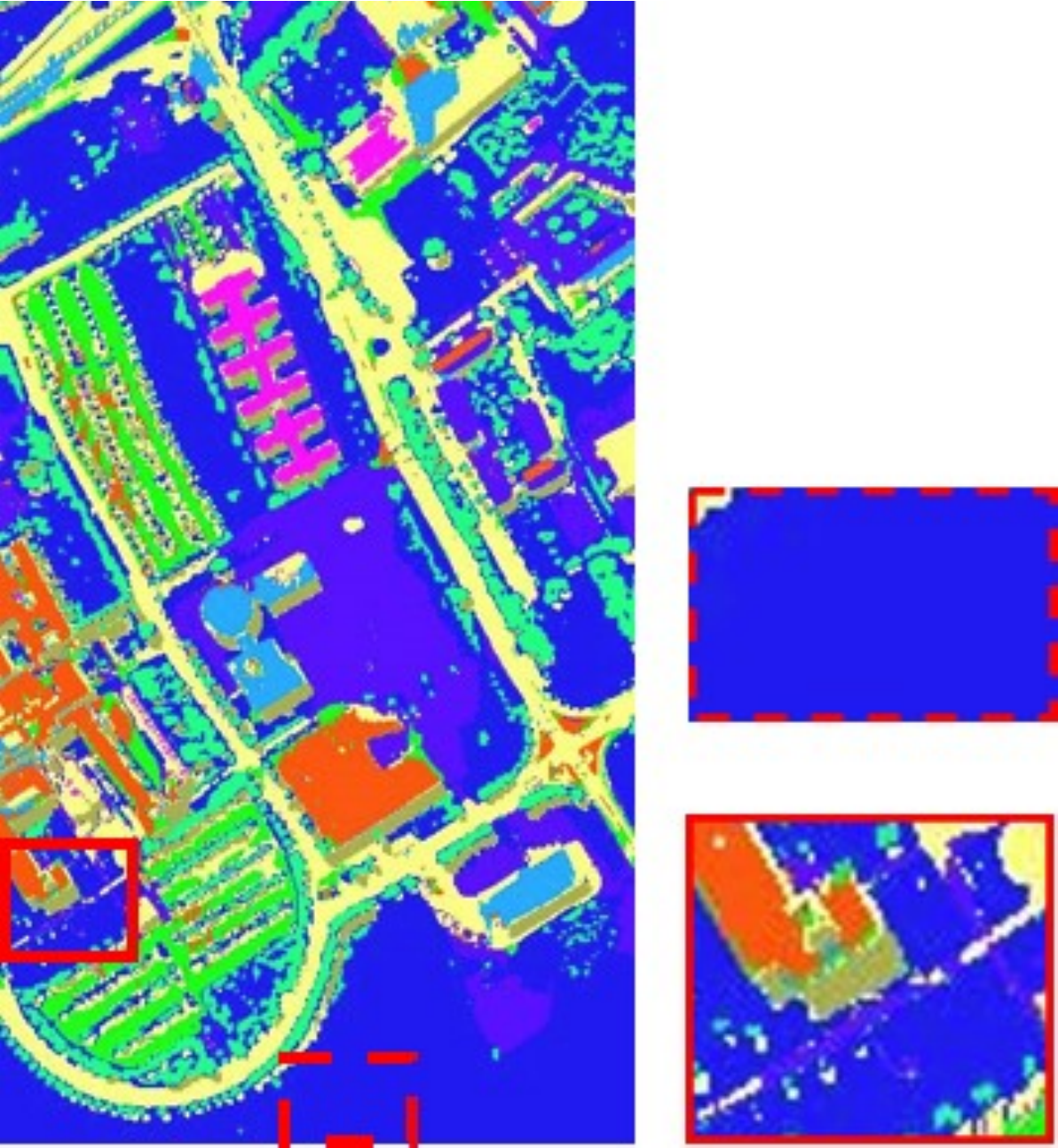}}\\
\centering \subfigure[]{
\includegraphics[width = 1.5in]{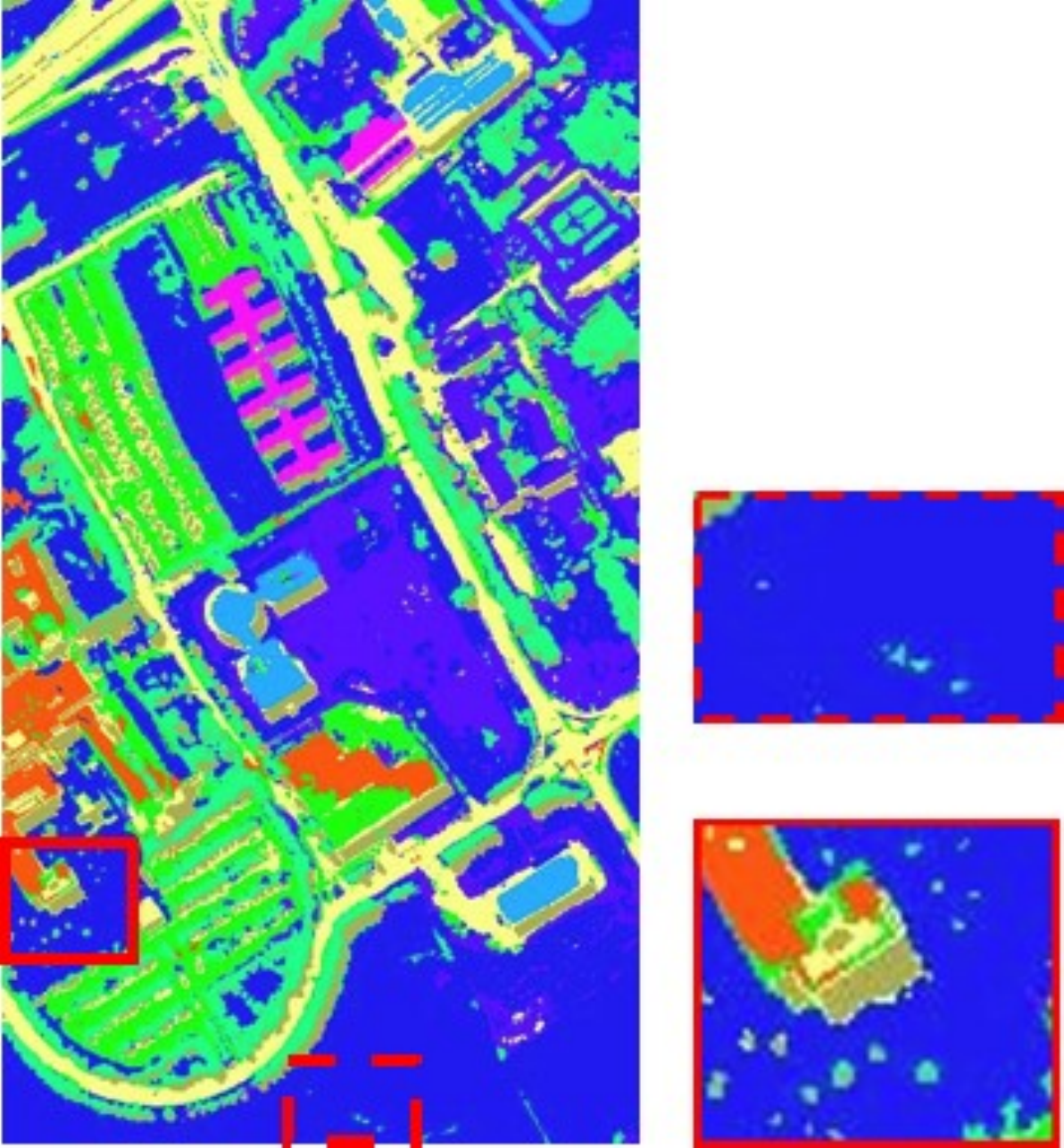}}
\centering \subfigure[]{
\includegraphics[width = 1.5in]{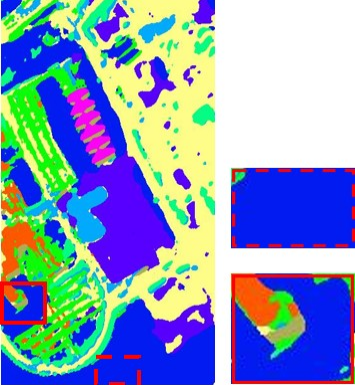}}
\centering \subfigure[]{
\includegraphics[width = 1.5in]{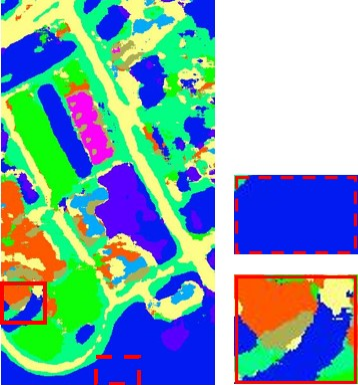}}\\
\centering \subfigure[]{
\includegraphics[width = 1.5in]{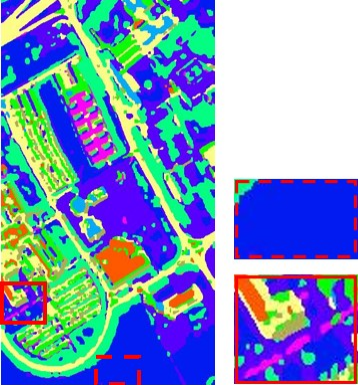}}
\centering \subfigure[]{
\includegraphics[width = 1.5in]{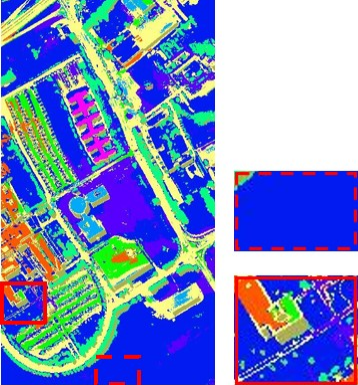}}
\centering \subfigure[]{
\includegraphics[width = 1.5in]{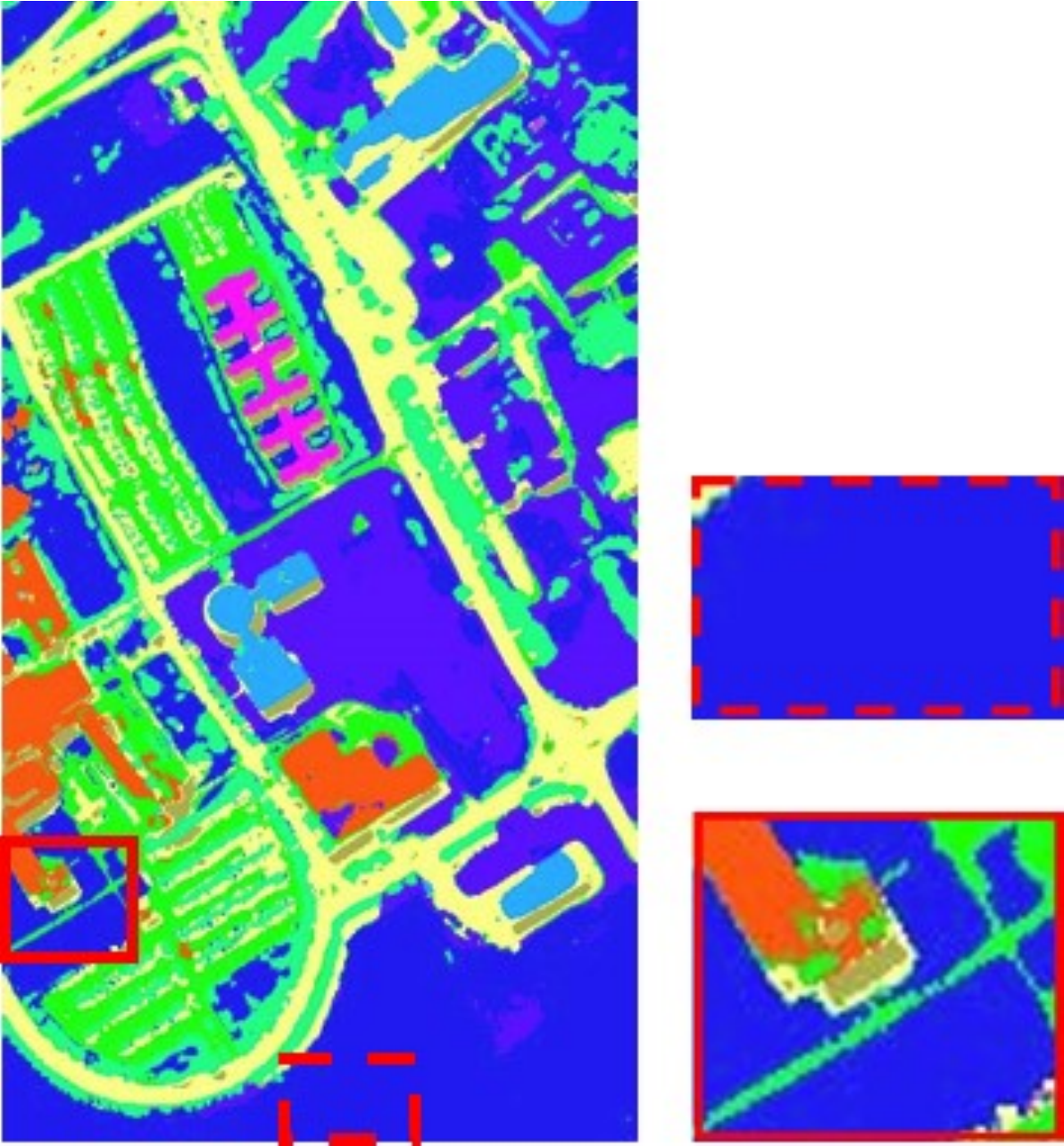}}
\caption{Classification maps on University of Pavia data set. (a) SADL; (b) MFL;  (c) PCA-EPF;  (d) HIFI;  (e) HybridSN; (f) SPDF; (g) CNN-MRF; (h) RPNet;  (i) MPRI.} \label{PaviaU_Map}
\end{figure}

\begin{figure}[htbp]
\centering \subfigure[]{
\includegraphics[width = 1.5in]{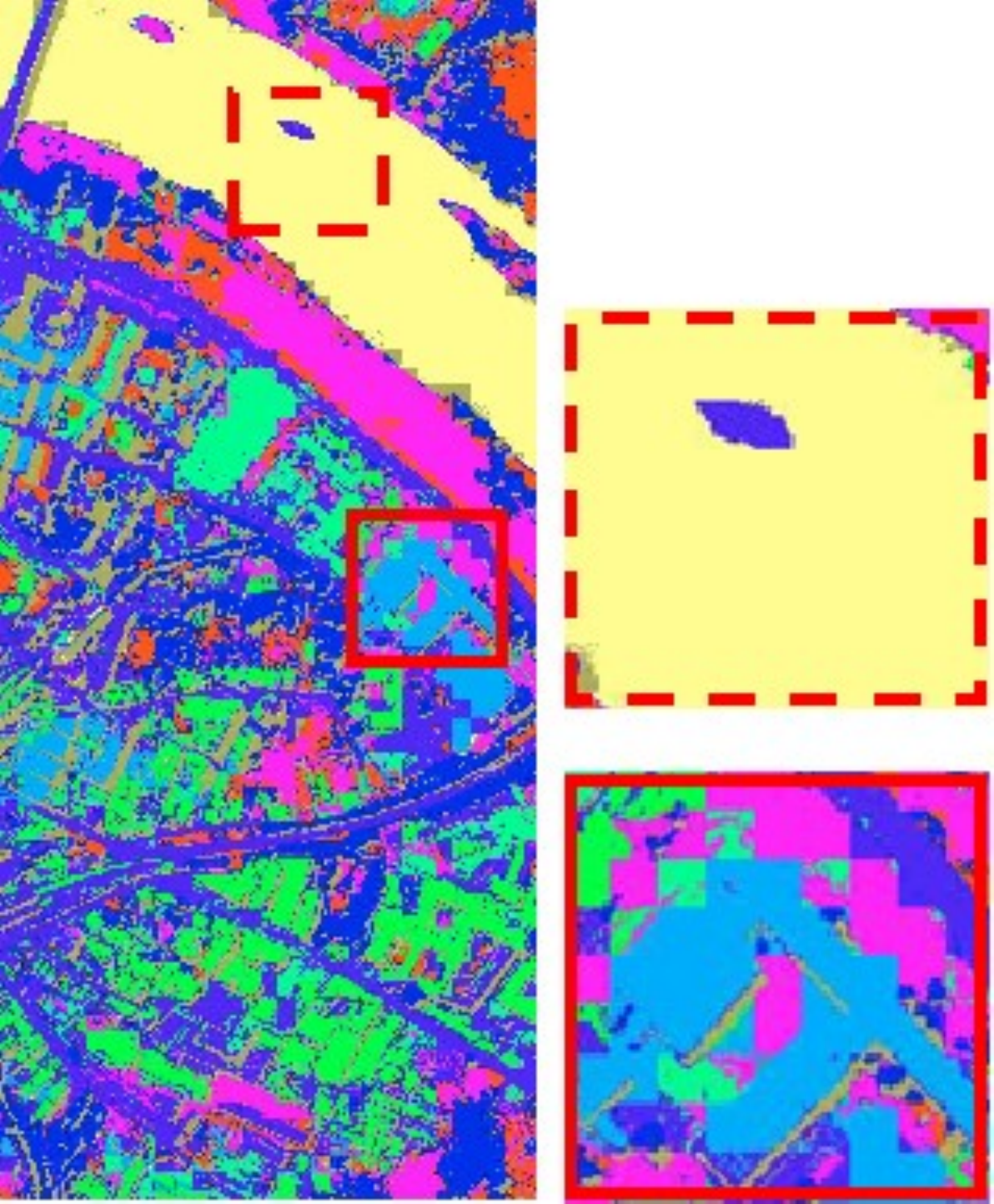}}
\centering \subfigure[]{
\includegraphics[width = 1.5in]{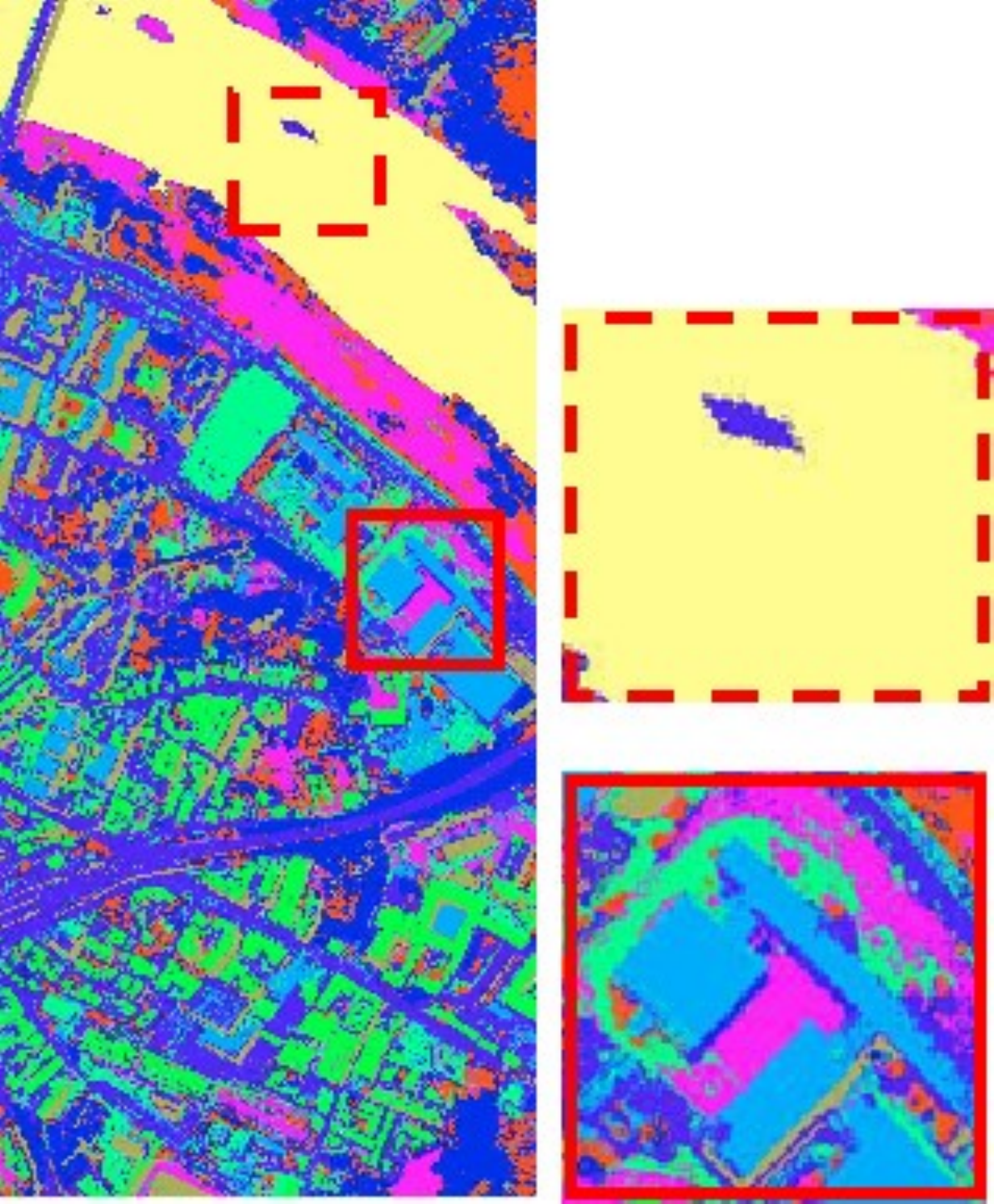}}
\centering \subfigure[]{
\includegraphics[width = 1.5in]{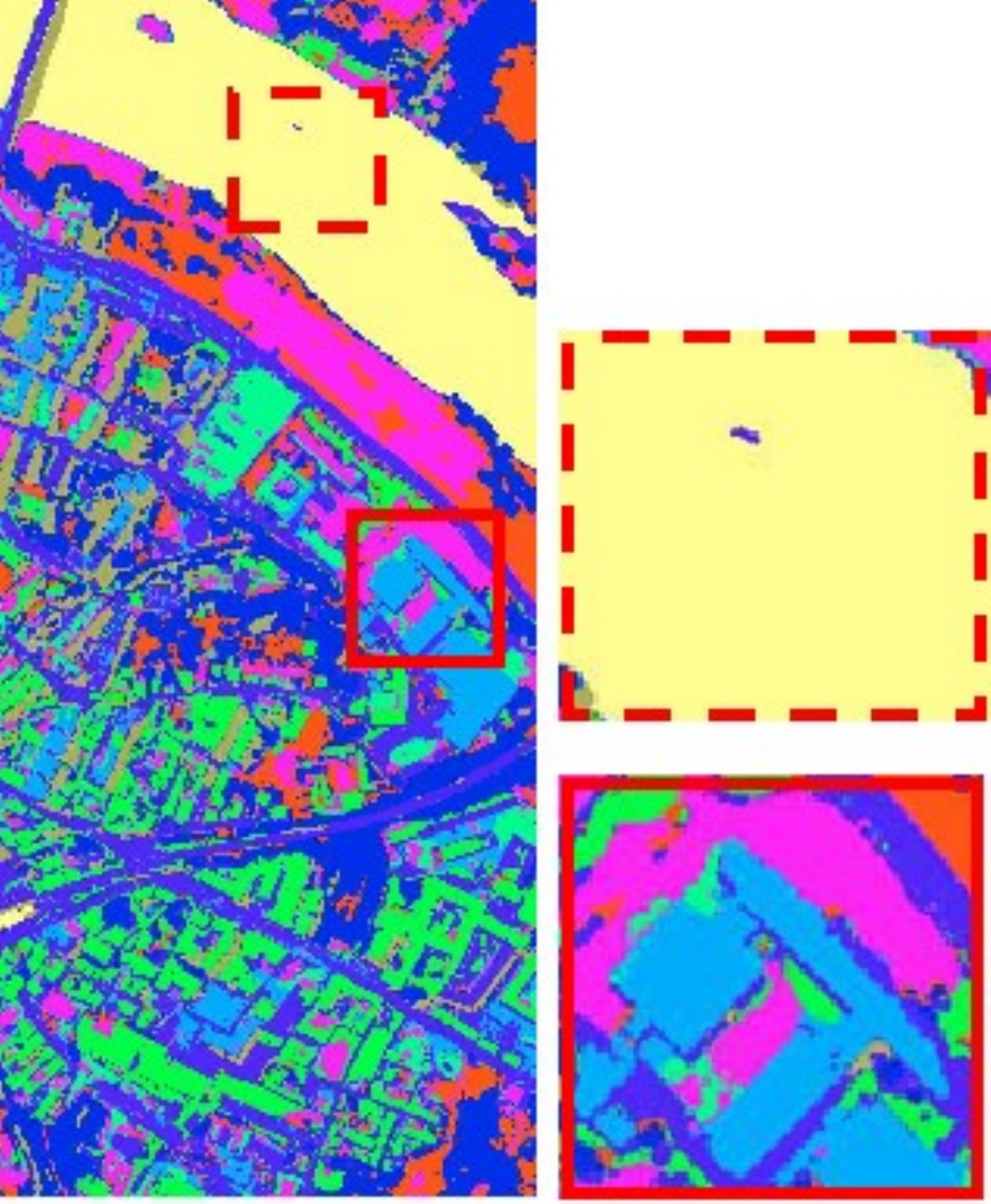}}\\
\centering \subfigure[]{
\includegraphics[width = 1.5in]{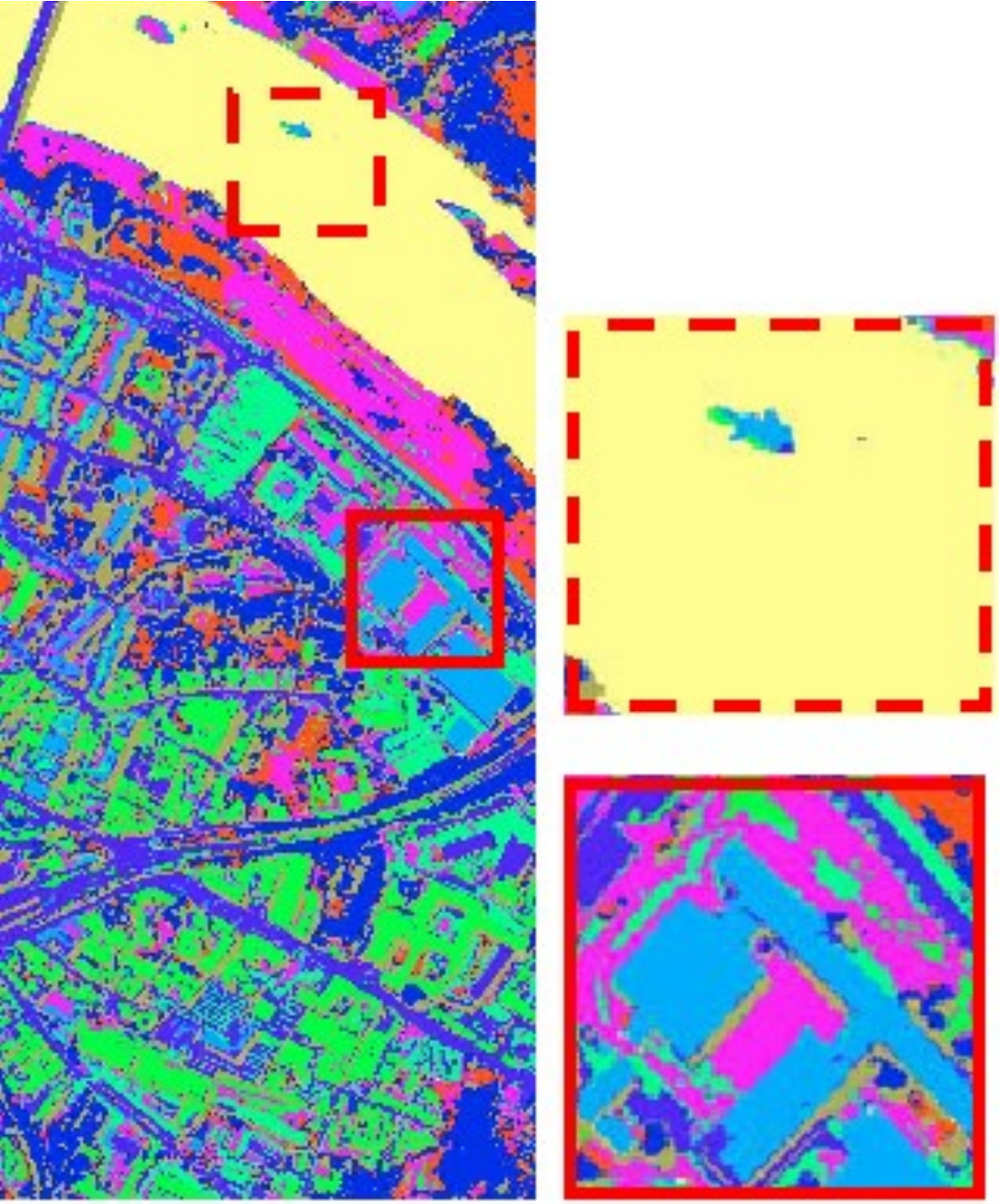}}
\centering \subfigure[]{
\includegraphics[width = 1.5in]{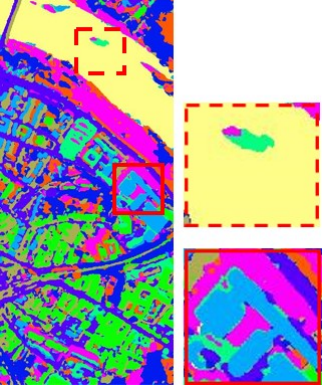}}
\centering \subfigure[]{
\includegraphics[width = 1.5in]{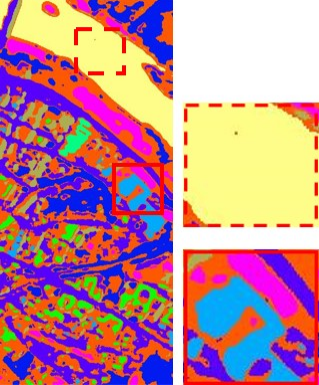}}\\
\centering \subfigure[]{
\includegraphics[width = 1.5in]{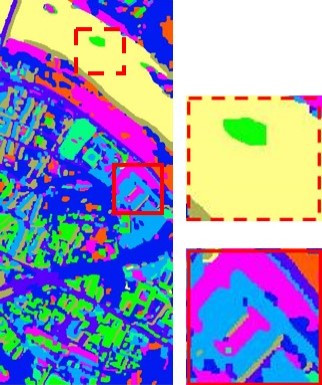}}
\centering \subfigure[]{
\includegraphics[width = 1.5in]{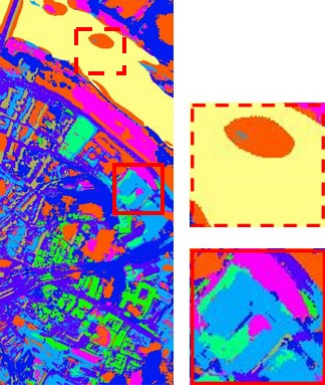}}
\centering \subfigure[]{
\includegraphics[width = 1.5in]{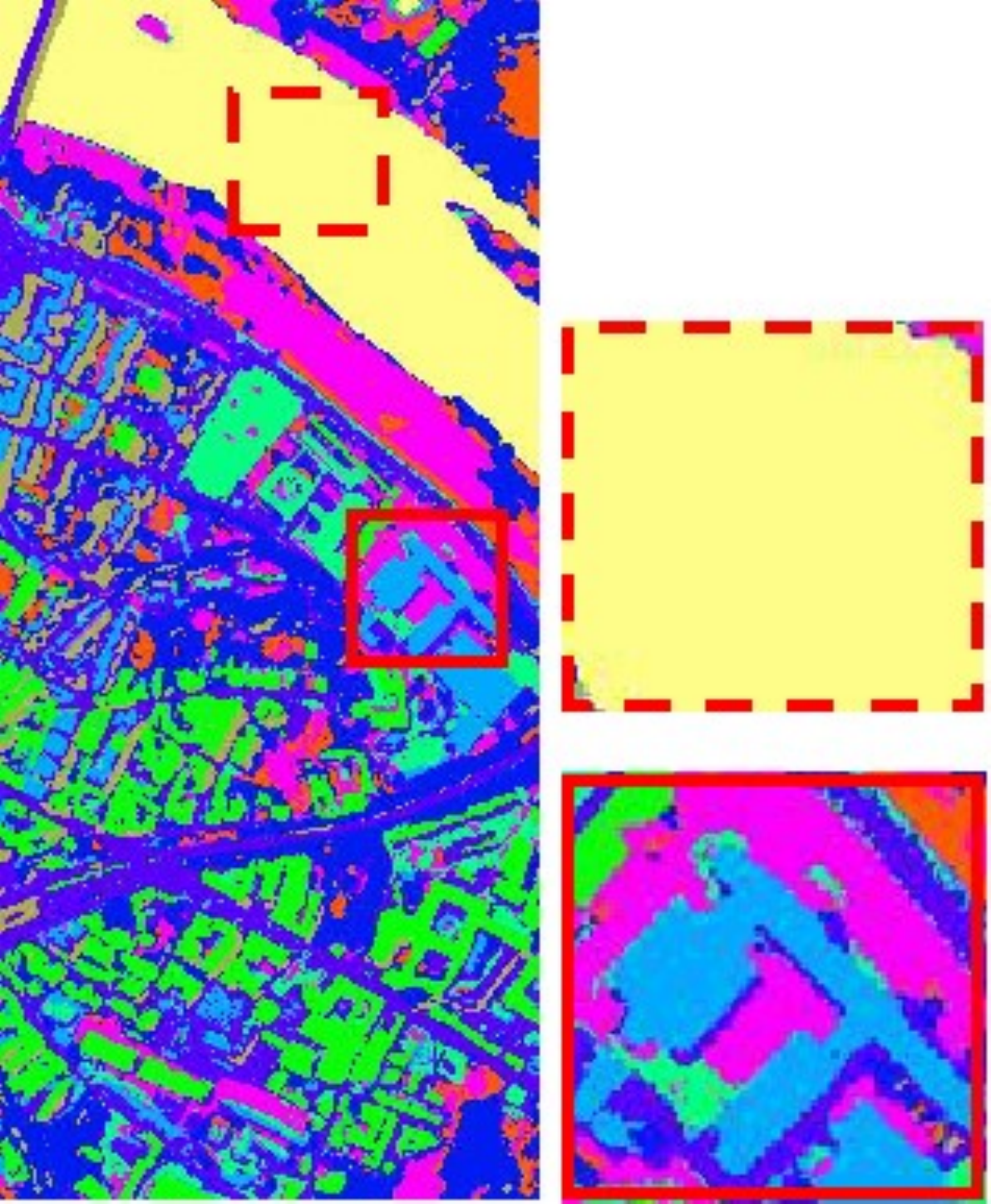}}
\caption{Classification maps on Pavia Center data set. (a) SADL; (b) MFL;  (c) PCA-EPF;  (d) HIFI;  (e) HybridSN; (f) SPDF; (g) CNN-MRF; (h) RPNet;  (i) MPRI.} \label{Pavia_Center_Map}
\end{figure}

The classification maps of different methods in three data sets are demonstrated in Figs.~\ref{Indian Pines_Train}-\ref{Pavia_Center_Map}, which further corroborate the above quantitative evaluations. The performances of EPF and MPM-LBP are omitted due to their relatively lower quantitative evaluations. It is very easy to observe that our proposed MPRI improves the region uniformity (see the small region marked with dashed border) and the edge preservation (see the small region marked by solid line rectangles) significantly, both criteria are critical for evaluating classification maps~\cite{kang2017pca}. By contrast, other methods either fail to preserve local details (such as edges) of different classes (e.g., MFL) or generate noises in the uniform regions (e.g., SADL, PCA-EPF and HIFI).

\begin{figure}[htbp]
\centering\subfigure[]{
\includegraphics[width = 4in]{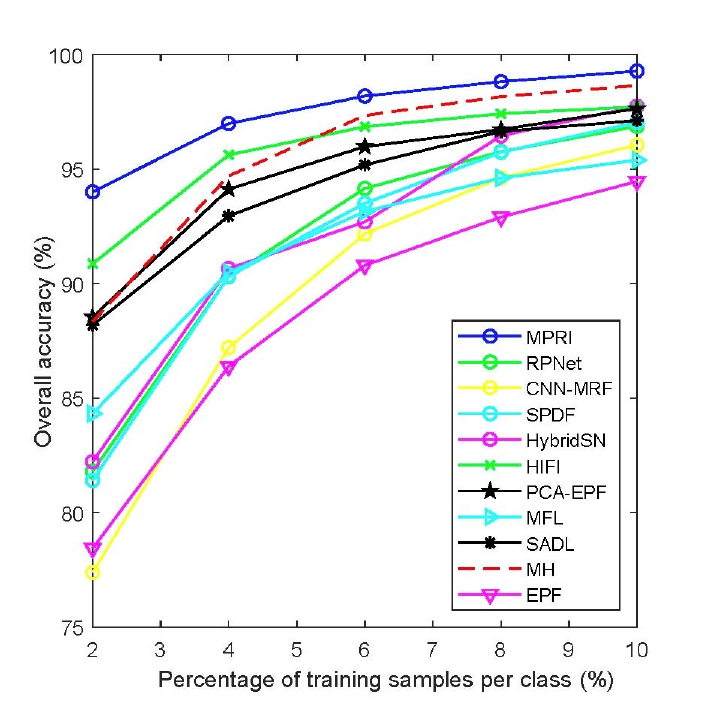}}
\centering\subfigure[]{
\includegraphics[width = 4in]{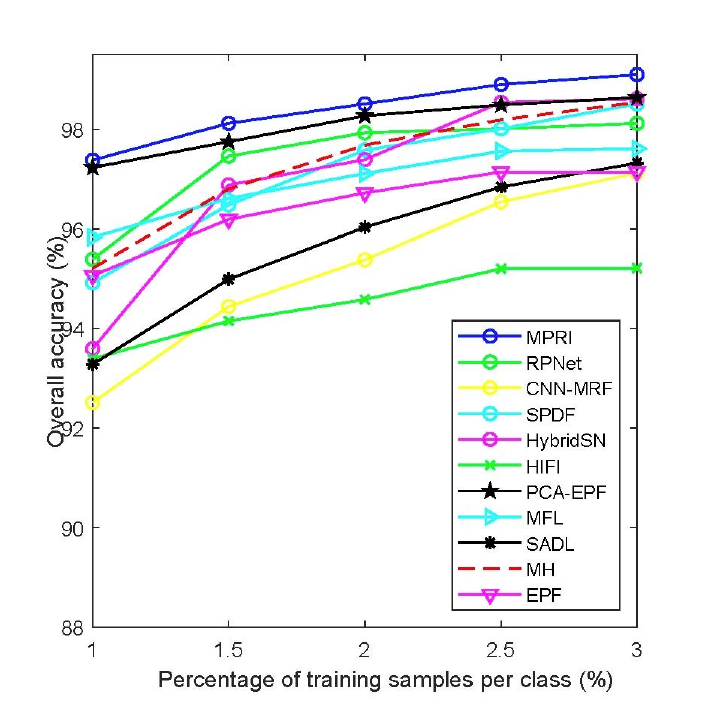}}
\caption{OA values of different methods with respect to different percentages of training samples per class on (a) Indian Pines; and (b) Pavia University. The results on Pavia  Center is omitted, because the training and testing samples are fixed.} \label{Indian_accuracy}
\end{figure}

To evaluate the robustness of our method with respect to the number of training samples, we demonstrate, in Fig.~\ref{Indian_accuracy}, the OA values of different methods in a range of the percentage of training samples per class. As can be expected, the more training samples, the better classification performance. However, MPRI is consistently superior to its counterparts, especially when the training samples are limited.

\subsection{Computational Complexity Analysis}


We finally investigate the computational complexity of different sliding window filtering based HSI classification methods. Note that, PRI can also be interpreted as a special kind of filtering, as the center pixel representation is determined by its surrounding pixels with a Gaussian weight (see Eq.~(\ref{xk})).

The computational complexity and the averaged running time on each pixel (in $s$) of different methods are summarized in Table~\ref{time}.
For PCA-EPF, ${\tilde d}$ is the dimension of averaged images, ${\hat S}$ is the number of different filter parameter settings, ${\hat T}$ is the number of iterations. For HIFI, $d$ is the number of hyperspectral bands, $\hat{H}$ is the number of the hierarchies. For MPRI, $L$, $S$, and $B$ are respectively the numbers of layers, scales and betas. Usually, ${\tilde d}$ is set to $16$, ${\hat S}$ is set to $3$, and ${\hat T}$ is set to $3$, which makes PCA-EPF very fast.

\begin{figure}[htbp]
\centering \subfigure[]{
\includegraphics[width = 1.5in]{Indian_MPRI.pdf}}
\centering \subfigure[]{
\includegraphics[width = 1.5in]{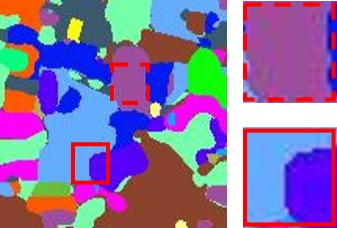}}
\caption{Classification maps of (a) MPRI and (b) MPRI with Nystr\"{o}m-KECA. The Gram matrix approximation is prone to cause over-smooth effect.} \label{Indian Pines_time}
\end{figure}

According to Eq.~(\ref{eq:PRI}), the computational complexity of PRI grows quadratically with data size (i.e., $\hat{N}$). Although one can simply apply rank deficient approximation to the Gram matrix for efficient computation of PRI, this strategy is preferable only when $l\ll\hat{N}$, where $l$ is the square of number of subsamples used to approximate the original Gram matrix~\cite{giraldo2011efficient}. In our application, $\hat{N}$ is less than a few hundreds ($\sim169$ at most), whereas we always need to set $l\geq25$ to guarantee a non-decreasing accuracy. From Table~\ref{time}, the reduced computational power by Gram matrix approximation is marginal. However, as shown in Fig.~\ref{Indian Pines_time}, such an approximation method is prone to cause over-smooth effect.

Finally, one should note that, although MPRI takes more time than its sliding window filtering based counterparts, it is still much more timesaving than prevalent DNN based methods.
For example, CNN-MRF takes more than $6,000s$ (on a PC equipped with a single $1080$ Ti GPU, i$7$ $8700$k CPU and $64$ GB RAM) to train a CNN model using $2\%$ labeled data on Indian pines data set with $10$x data augmentation.





\begin{table}[htbp]
\caption{Computation complexity and running time (per pixel) of different methods.}
\begin{center}{\begin{tabular}{lcc}
\toprule
Method & Computational complexity & Running time per pixel ($s$)\\
\midrule
EPF& $\mathcal{O}(1)$& $2.9 \times 10^{-6}$\\
PCA-EPF&$\mathcal{O}({\tilde d}{\hat S}{\hat T})$&$2.6\times 10^{-5}$\\
HIFI&$\mathcal{O}(d{\hat{H}})$& $1.0 \times 10^{-3}$\\
MPRI & $\mathcal{O}(d\hat{N}^2 \tau LSB)$  &$3.2\times 10^{-1}$\\
MPRI with Nystr\"{o}m-KECA &$\mathcal{O}(d{\hat N} l \tau LSB)$ &$2.6\times10^{-1}$\\
\bottomrule
\end{tabular}}
\end{center}
\label{time}
\end{table}

\section{Conclusions}\label{chapter_conclusion}

This paper proposes multiscale principle of relevant information (MPRI) for hyperspectral image (HSI) classification. MPRI uses PRI - an unsupervised information-theoretic learning principle that aims to perform mode decomposition of a random variable $X$ with a known (and fixed) probability distribution $g$ by a hyperparameter $\beta$  - as the basic building block. It integrates multiple such blocks into a multiscale (by using sliding windows of different sizes) and multilayer (by stacking PRI successively) structure to extract spectral-spatial features of HSI data from a coarse-to-fine manner. Different from existing deep neural networks, MPRI can be efficiently trained greedy layer-wisely without error backpropagation. Empirical evidence indicates $\beta\in[2,4]$ in PRI is able to balance the trade-off between the regularity of extracted representation and its discriminative power to HSI data. Comparative studies on three benchmark data sets demonstrate that MPRI is able to learn discriminative representations from $3$D spatial-spectral data, with significantly fewer training samples. Moreover, MPRI enjoys an intuitive geometric interpretation, it also prompts the region uniformity and edge preservation of classification maps. In the future, we intend to speed up the optimization of PRI. In this line of research, the random fourier feature~\cite{rahimi2008random} seems to be a promising avenue.



\begin{acknowledgements}
This research was funded by the National Natural Science Foundation of China under Grants 61502195, the Humanities and Social Sciences Foundation of the Ministry of Education under Grant 20YJC880100, the Natural Science Foundation of Hubei Province under Grant 2018CFB691, the Fundamental Research Funds for the Central Universities under Grant CCNU20TD005, and the Open Fund of Hubei Research Center for Educational Informationization, Central China Normal University under Grant HRCEI2020F0101.
\end{acknowledgements}

\bibliographystyle{spmpsci}
\bibliography{IEEEabrv_N}

\end{document}